\documentclass[journal]{IEEEtran}
\ifCLASSINFOpdf
  \usepackage[pdftex]{graphicx}
\else
\fi

\usepackage{mathrsfs}
\usepackage{multirow}
\usepackage{booktabs}
\usepackage{hyperref}
\usepackage{color}
\usepackage{amsmath, bm}
\usepackage[ruled,linesnumbered]{algorithm2e}
\usepackage{tabu}
\usepackage{cancel}
\usepackage{subfig}
\usepackage[table]{xcolor}
\usepackage{fp}  % Add this in your preamble

\usepackage{flushend}

\usepackage{mathrsfs}

\ifCLASSINFOpdf
\else
\fi
\usepackage{amssymb}
\usepackage{mathrsfs}
\usepackage{amsmath}
\usepackage{bm}
\usepackage{makecell}
\usepackage{multirow}
\usepackage{colortbl}
\usepackage{soul}
\usepackage{caption}
\usepackage[figuresleft]{rotating}
\definecolor{chromeyellow}{rgb}{1.0, 0.65, 0.0}
\definecolor{carmine}{rgb}{0.59, 0.0, 0.09}
\usepackage[percent]{overpic}
\def\eg{\emph{e.g.}}
\def\ie{\emph{i.e.}}
\def\etal{\emph{et al. }}

\begin{document}

\captionsetup[table]{skip=0pt}
% \title{Image Deblurring via In-depth Transformer Properties}
\title{Image Deblurring by Exploring In-depth Properties of Transformer}
% \title{Elevating Image Deblurring with Transformer Properties}

\author{Pengwei Liang,
        Junjun~Jiang,~\IEEEmembership{Senior Member,~IEEE,~}
        Xianming Liu,~\IEEEmembership{Member,~IEEE,~}
        and Jiayi Ma,~\IEEEmembership{Senior Member,~IEEE}% <-this % stops a space

 \thanks{The research was supported by the National Natural Science Foundation of China (U23B2009, 92270116, 62276192).(Corresponding author: Junjun Jiang)
 } %
 
% \thanks{Manuscript received January 24, 2022. revised February 15, 2022; accepted March 3, 2022. \emph{(Corresponding author: Junjun Jiang)}}
\thanks{P. Liang, J. Jiang and X. Liu are with the School of Computer Science and Technology, Harbin Institute of Technology, Harbin 150001. E-mail: \{jiangjunjun, csxm\}@hit.edu.cn.}% <-this % stops a space
%\thanks{J. Jiang and X. Liu are with the School of Computer Science and Technology, Harbin Institute of Technology, Harbin 150001, China. E-mail: \{jiangjunjun, csxm\}@hit.edu.cn.}% <-this % stops a space
\thanks{J. Ma is with the Electronic Information School, Wuhan University, Wuhan 430072, China. E-mail: jyma2010@gmail.com.}
}

\markboth{Journal of \LaTeX\ Class Files,~Vol.~14, No.~8, August~2021}%
{Shell \MakeLowercase{\textit{et al.}}: Bare Demo of IEEEtran.cls for IEEE Journals}



\maketitle

\begin{abstract}
Image deblurring continues to achieve impressive performance with the development of generative models. Nonetheless, there still remains a displeasing problem if one wants to improve perceptual quality and quantitative scores of recovered image at the same time. In this study, drawing inspiration from the research of transformer properties, we introduce the pretrained transformers to address this problem. In particular, we leverage deep features extracted from a pretrained vision transformer (ViT) to encourage recovered images to be sharp without sacrificing the performance measured by the quantitative metrics. The pretrained transformer can capture the global topological relations (\ie, self-similarity) of image, and we observe that the captured topological relations about the sharp image will change when blur occurs. By comparing the transformer features between recovered image and target one, the pretrained transformer provides high-resolution blur-sensitive semantic information, which is critical in measuring the sharpness of the deblurred image. On the basis of the advantages, we present two types of novel perceptual losses to guide image deblurring. One regards the features as vectors and computes the discrepancy between representations extracted from recovered image and target one in Euclidean space. The other type considers the features extracted from an image as a distribution and compares the distribution discrepancy between recovered image and target one. We demonstrate the effectiveness of transformer properties in improving the perceptual quality while not sacrificing the quantitative scores (PSNR) over the most competitive models, such as Uformer, Restormer, and NAFNet, on defocus deblurring and motion deblurring tasks. The code is available at https://github.com/erfect2020/TransformerPerceptualLoss.
\end{abstract}

% By comparing the topological relationships between blurred and clear images, the pre-trained transformer provides high-resolution blur-sensitive semantic information, which is essential for measuring the sharpness of deblurred images.
% In particular, we use deep features extracted from the pre-trained visual transformer (ViT) to encourage clarity in the recovered images without sacrificing the performance measured by the PSNR metric.
% One proposed perceptual loss regards the features as vectors and computes the distance between representations extracted from blurred and sharp image in Euclidean space.
% A proposed perceptual loss treats features as vectors and computes the distance between representations extracted from blurred and clear images in Euclidean space.

% \fi
\IEEEpeerreviewmaketitle

\section{Introduction}\label{sec1}

\begin{figure}[htbp]
    \centering
    \includegraphics[width=8.8cm]{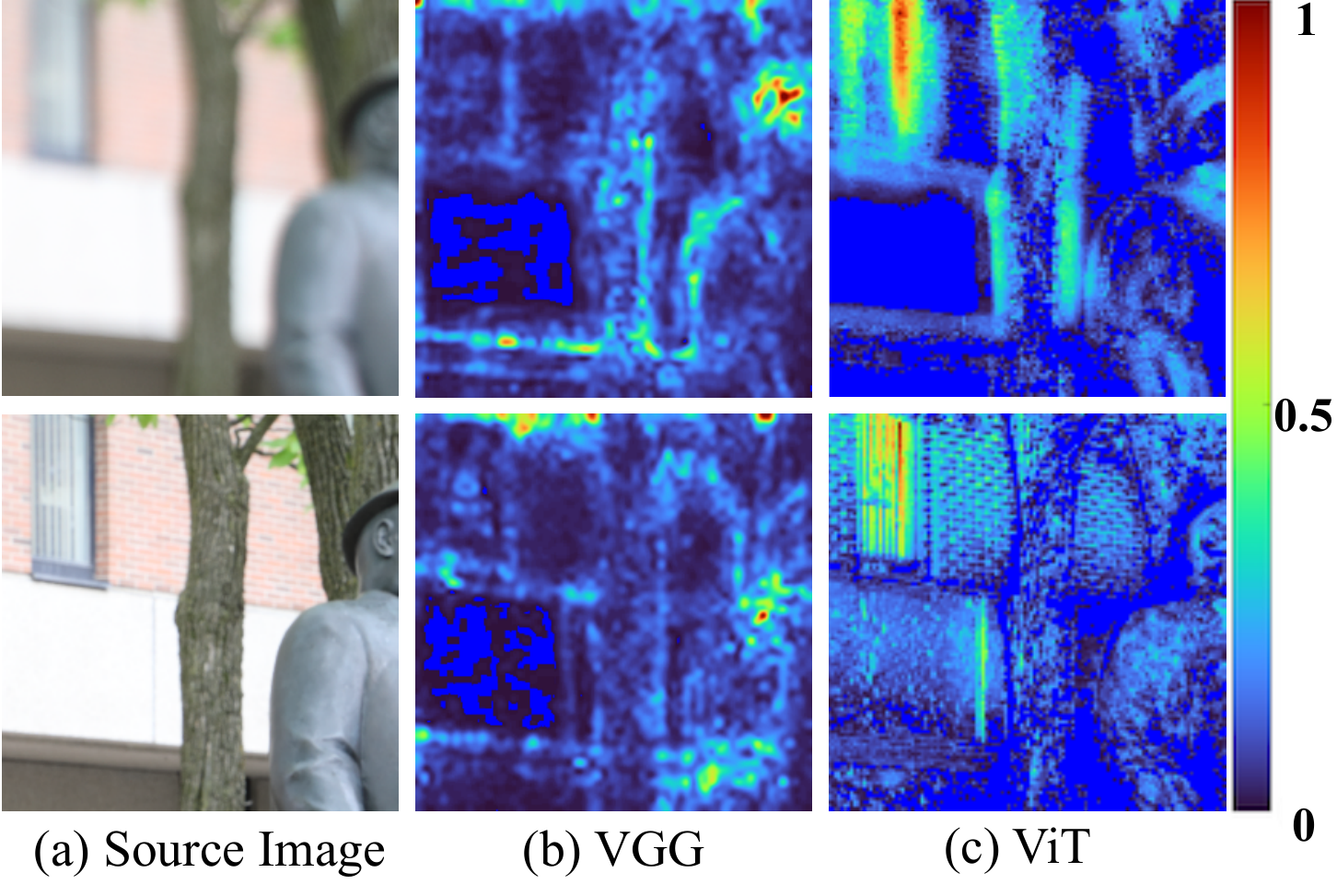}
    \caption{Illustration the effectiveness of VGG and ViT features. The first/second row presents a blurry/ground truth example and their feature representations. By comparing the visualized features of the last column, we can observe that the image blur obviously changes the topological relations (\ie, self-similarity) of features extracted from the clean images. % We employ the turbo colormap to transform the grayscale image into a pseudo-color image.
    }
    \vspace{-10pt}
    % By comparing the visualized features of last column, we can observe that the blurry change the topological relations of features extracted from sharp images.
    \label{fig:first intorduction}
\end{figure}

% Image deblurring is a classic task in low-level computer vision, which has attracted the attention from the image processing and computer vision community. The objective of image deblurring is to recover a sharp image from a blurred input image, where the blur can be caused by various factors such as lack of focus, camera shake, or fast target motion
\IEEEPARstart{I}{mage} deblurring is a long-standing problem in low-level computer vision, which aims to recover a sharp image from a blurred input \cite{zhang2022deep}. There are various factors that can lead to the blurry. For instance, defocus blurring arises when depth variation goes beyond the allowable range of focus \cite{shi2014discriminative}; motion blur is caused by camera shakes or object motions during the exposure in dynamic environments \cite{hyun2015generalized}. The complicated degradation process makes it very challenging to accurately estimate the blur kernel by formulating the image deblurring as an inverse filtering kernel problem. To avoid explicitly estimating the blur kernel, it has become popular for deblurring methods to directly recover the sharp image from the blur image via an end-to-end learning way. Recently, a variety of end-to-end methods have been proposed that have achieved promising performance. It is worth noting that most of these deblurring studies use pixel-wise loss functions such as $\ell_1$ or $\ell_2$ to minimize the distance between recovered images and reference images \cite{abuolaim2020defocus,wang2022uformer, zamir2022restormer, chen2022simple, liang2022bambnet}.

% as major training objective []. 

However, the pixel-wise losses will suffer from some limitations when handling the image blind deblurring tasks. Most commonly, the losses of $\ell_1$ and $\ell_2$ tend to drive the model to produce a blurred prediction with poor perceptual quality. To address this deficiency, Johnson \etal \cite{johnson2016perceptual} introduced a perceptual loss (notated as the classical perceptual loss) as an additional regularization term into the image restoration optimization. %The classical perceptual loss uses high-level features extracted from the pretrained VGG network to capture perceptual differences between the reconstructed image and ground truth. 
However, the improvement of visual quality is always accompanied by a decrease in quantitative metrics (\eg, peak signal-to-noise ratio (PSNR) ) \cite{nah2021clean,delbracio2021projected}. This has been avowed by many low-level vision tasks \cite{ledig2017photo,kupyn2019deblurgan,zhang2022deep}.

% (
% T)he other is that the weight of perceptual regularization term requires to be finetuned very carefully to tradeoff between perception and distortion.
% Improvement in visual quality is always accompanied by a decrease in the quality fraction of the reconstructed image

In this paper, we thoroughly study how to improve the perceptual quality of the deblurred image without sacrificing the quantitative scores. Existing methods usually adopt the pretrained VGG as the off-the-shelf feature extractor, while the extracted feature maps are of very low resolution. In particular, Zhang \etal \cite{zhang2020deblurring} calculated the classical perceptual loss using features before the pool5 layer of VGG for motion deblurring. The size of feature maps is 32$\times$ smaller than taht of the source input. Such high-level feature maps with low spatial resolution inevitably miss rich details and provide poorly localized global semantic information \cite{amir2021deep}, as shown in Fig. \ref{fig:first intorduction}. Moreover, the coarse feature maps from VGG might work well in abstract high-level tasks, but they may not be helpful in guiding image restoration in low-level vision tasks. To better assist  the image deblur task, we introduce the pretrained transformer to replace the CNN-based network due to the following three factors. First, due to the global self-attention mechanism, the transformer can well model the self-similarity of natural images, and this is very important for the low-level image restoration task \cite{chen2021pre}. Second, unlike VGG where the resolution of the feature map gradually decreases as layers go deeper, the vision transformer (ViT) \cite{dosovitskiy2020vit} maintains the same spatial resolution through all layers. As shown in Fig. \ref{fig:first intorduction}(c), the visualization of the ViT features from the last layer still reveals abundant details of the image. By close comparison of the visualizations between blurred and sharp images, it can be observed that the features produced by VGG exhibit only slight disparities, while those generated from the ViT show significant differences. % It indicates that the features from the ViT are more sensitive to blurriness. 
Third, recent works \cite{amir2021deep, tumanyan2022splicing} reveal that even representations extracted from the last layer of ViT can still reconstruct the original image. It indicates that the features extracted from ViT preserve significantly fine-grained information in each layer. 

% quantitative results measured by peak signal-to-noise ratio (PSNR) and structural similarity index (SSIM)

% It proves that the properties of ViT preserve rich information in each layer.
% such that moderately shufﬂing the elements does not degrade the performance signiﬁcantly.
% Owing to the low spatial resolution, high-level feature maps inevitably miss voluminous details.
% Unlike VGG where the feature map resolution decreases as the layers go deeper, the Visual Transformer (ViT) maintains the same spatial resolution in all layers.

% In the last type, we compute the distance of similarity maps that capture the global interaction and collaboration of transformer features as shown in Fig. \ref{fig:similarity map}(b-d).  

% To improve the quality of deblurred images,

\begin{figure}[t]
    \centering
    \includegraphics[width=9cm]{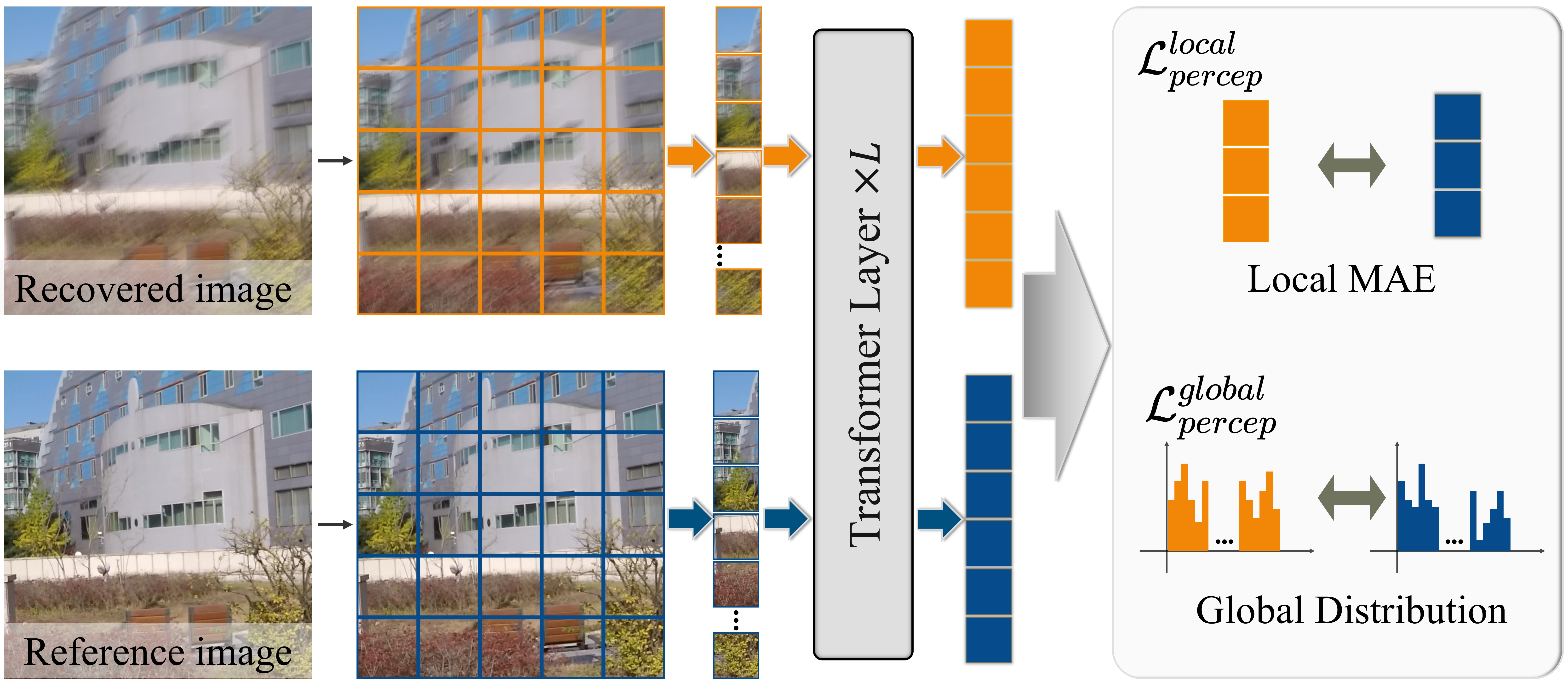}
    \caption{The workflow illustration of proposed perceptual losses.}
    \vspace{-10pt}
    \label{fig:objectives}
\end{figure}

Motivated by above observations, we design two types of perceptual losses with the features extracted from the pretrained transformer from the local and global perspectives (as shown in Fig. \ref{fig:objectives}).
% using the pretrained transformer as regularization terms of the pixel-wise loss function. 
% Regularization terms penalize the disparity between the deblurred and target representations extracted from pretrained transformer. 
Precisely, for the first type, we consider the element-wise features as vectors in Euclidean space and then directly compute their differences. For the second type, instead of aligning the element-wise feature at each position, we compare the global distribution of features based on the Optimal Transport (OT) theory \cite{villani2009optimal}. The first loss focuses on aligning the local feature information, which may facilitate to generate deblurred image with few artifacts. While the strict constraint may limit the deblurred image to generate more details. The latter compares the distribution of features, which may produce more realistic images. % However, the recovered details with global perceptual loss may not be located at the exact same spatial position as in the reference image.

% s penalizes difference in the statistics of the feature maps

% By comparing the distribution of features in addition to the pixel values, we are able to produce images with a higher level of realism.
% This allows recovering details (texture/grain) without forcing them to be located in the exact same spatial position as in the reference image.

The main contribution of this work is incorporating transformer properties into image deblurring task, which improves the perceptual quality of the reconstructed image, while rarely arising performance drop in quantitative metrics (\eg, PSNR). We carefully investigate how to design novel perceptual losses to guide image deblurring. Extensive experiments demonstrate that the proposed perceptual losses can effectively boost the performance when applied to the most competitive approaches. Furthermore, we show that the proposed perceptual losses are general can be easily combined with other losses.

% , where the weight of semantic loss is robust to scale
% version 2
% Real image blurring exists significant locally-varying blur characteristics domained by the complicated degradation. For instance, depth variation might cause the defoucs blurring; camera shakes or object motions in dynamic environments results in motion blur.

\section{Related Work}\label{sec2}

\subsection{Image Deblurring}

% According to the nature of blur, image blur can be classified as out-of-focus blur and motion blur.
According to the nature of blur, image blur can be classified into defocus blur and motion blur \cite{zhang2022deep}. In the deep learning era, a variety of deblurring methods have been proposed to address the problem. To be specific, Abuolaim and Brown \cite{abuolaim2020defocus} provided a large-scale dataset, called DPDD, for the defocus deblurring task. In addition to the standard benchmark dataset, they also proposed a simple U-Net-like architecture as a baseline. At a later time, they further introduced a multi-decoder network architecture to improve the performance of deblurring \cite{abdullah2021rdpd}. Recently, more and more powerful network modules have been designed for defocus deblurring, including transformer block \cite{wang2022uformer, zamir2022restormer}, image representation \cite{xin2021defocus}, iterative filter adaptive module \cite{Lee_2021_CVPRifan}, and kernel design \cite{son2021single, ruan2022learning, gong2020learning}. Besides, since the defocus blur is related to the depth variation, some works estimate the defocus map based on physical rules to guide the network to handle the non-uniform blur \cite{pan2021dual,liang2022bambnet,ma2021defocus}. Compared to the defocus blur, motion blur has attracted more widespread attention in the early days. Specifically, Nah \etal \cite{nah2017deep} first proposed a DeepDeblur that directly maps the blurry image into its sharp counterpart without explicit kernel estimation in 2017. Several novel techniques have been successfully adopted in motion deblurring, such as multi-scale network \cite{nah2017deep, liu2021multi}, scale-recurrent architectures \cite{tao2018scale, gao2019dynamic}, adversarial training \cite{kupyn2018deblurgan, kupyn2019deblurganv2, wen2021structure, zhang2022blind}, spatial frequency domain \cite{mao2021deep}, reblurring \cite{nah2021clean, zhang2020deblurring, bahat2017non}, event sensor \cite{chen2022residual}, and coarse-to-fine strategies \cite{cho2021rethinking_mimo}. Nevertheless, those methods focus on designing the network architectures and rarely explore improving image quality in terms of quantitative and perceptual measurements by designing novel losses.

\subsection{Perceptual Loss in Image Restoration}

The classical loss is introduced by Johnson \etal \cite{johnson2016perceptual} for style transfer and super-resolution. The classical perceptual loss depends on high-level features extracted from pretrained VGG networks to capture latent semantic knowledge. By training with classical perceptual loss, the model can produce more visually pleasing results. Then, perceptual loss became popular in the computer vision community and spread into numerous image restoration tasks, such as image reflection removal \cite{zhang2018single}, image dehazing \cite{zhang2018multi}, and image deblurring \cite{zhang2020deblurring}. Inspired by the classical perceptual loss, Zhang \etal \cite{zhang2018unreasonable} further conducted broad experiments and revealed that intermediate deep features can model low-level perceptual similarity surprisingly well. Based on experimental observations, they proposed an image quality assessment metric, \ie, learned perceptual image patch similarity (LPIPS). In addition to computing the element-wise distance of extracted feature maps, Delbracio \etal \cite{delbracio2021projected} proposed to compare feature distributions between predicted and target images by projecting the multidimensional feature space into one dimension.

\begin{figure}[t]
    \centering
    \includegraphics[width=8.8cm]{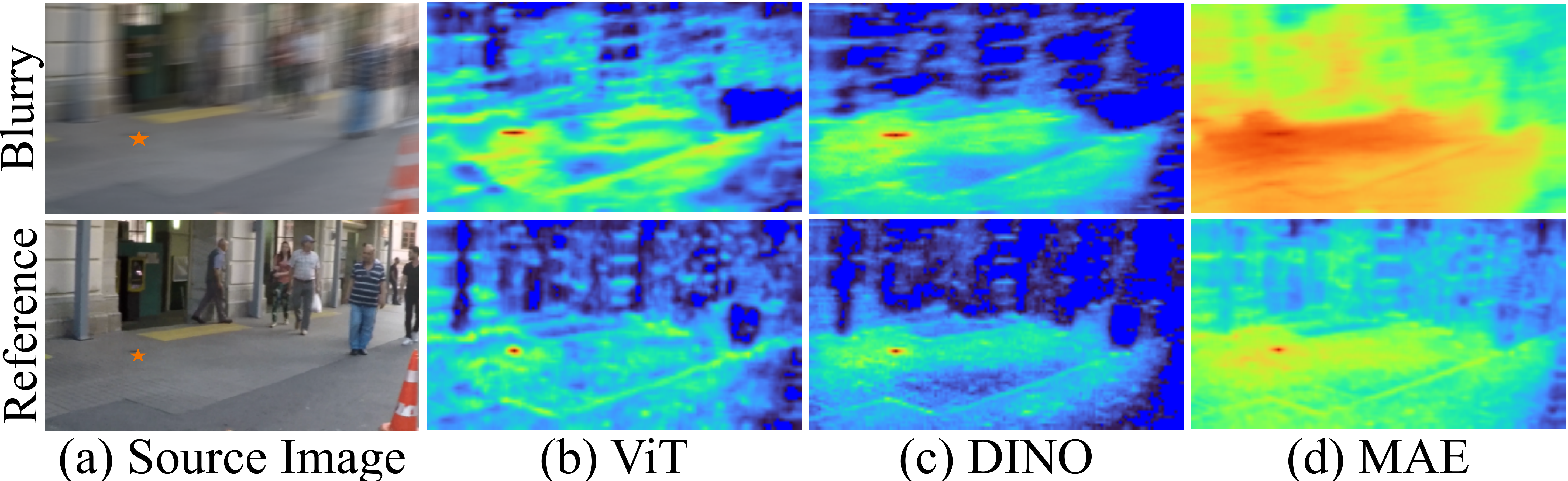}
    \caption{Similarity visualization of different ViT features on the image deblurring dataset. We extract the ViT features from the same ViT architecture training in different ways: (b) supervised ViT \cite{dosovitskiy2020vit}, (c) self-supervised DINO \cite{caron2021emerging}, and (d) MAE \cite{MaskedAutoencoders2021}. The (b-d) show similarity heatmaps that are computed between a feature located at $\color{chromeyellow}\star$ and all features in this image. %The closer the color is to red, the higher the similarity.
    }
    \vspace{-10pt}
    \label{fig:similarity map}
\end{figure}

However, some work found that the performance of the quantitative measure (\eg, PSNR) decreases when the classical perceptual loss is introduced to improve the perceptual quality of recovered images \cite{zhang2022deep,ledig2017photo, kupyn2019deblurgan}. Therefore, the authors have to tune the weight of perceptual loss very carefully to maximize perception gain and minimize the decrease in quantitative scores. In this paper, we address this limitation using transformer feature representations extracted from pretrained masked autoencoder (MAE) \cite{MaskedAutoencoders2021}. Different from the pretrained VGG, the pretrained transformer shows surprising properties in which intermediate features preserve almost all details and exhibit high-resolution blur-sensitive information. Next, we simply review the vision transformer (ViT) properties revealed in high-level vision tasks.
% Experiments demonstrate that the transformer features are able to better guide the image deblurring where human perception score is improved while the performance of distortion measure does not decrease.

\subsection{Transformer Properties in High-level Vision Tasks}

As ViT has shown impressive performance across numerous high-level vision tasks, it attracts more attention to intrigue properties of ViT \cite{liu2023survey}. Various works based on the ViT have achieved impressive performance in the computer vision community \cite{ma2023rectify, tian2023end, li2023bvit, du2023boosting, xu2023multi}. Naseer \etal \cite{naseer2021intriguing} presented that supervised ViT shows promising robustness properties in occlusions, adversarial attacks, and shape recognition. They also found that off-the-shelf supervised ViT features exhibit strong generalizability to new classification and recognition domains. Following this work, Amir \etal \cite{amir2021deep} explored the properties of a self-supervised ViT (DINO-ViT). They demonstrated that the DINO-ViT (abbreviated as DINO in the following sections) has learned more powerful visual representations encoding high spatial resolution semantic information compared to supervised ViT and pretrained CNN. Then, the authors extract the deep features from DINO as dense ViT descriptors that are applied to image retrieval, object segmentation, and copy detection applications. Even using only the ViT descriptors, they achieved competitive results on those tasks without any additional training or fine-tuning. In addition, Cao \etal \cite{cao2022understand} proved that each token in MAE has learned global inter-token topological relations using rigorous mathematical theory.

% Moreover, the kernel map is globally supported, which means it can learn effectively the interaction between even far away patches

% We prove that the latent representations of the masked patches are interpolated globally based on an inter-patch topology that is learned by the attention mechanism.

However, the above-mentioned studies pay more attention to high-level semantic representation understanding. It is not determined whether the abstracted representations of ViTs are useful for low-level tasks or not. In this paper, we are an effort to explore the effectiveness of ViT properties on low-level image restoration tasks.

\section{Method}\label{sec3}

%In this section, we first show an observation revealing that transformer features can be used to indicate blurring. Next, we briefly review the overview of ViT architecture, and then propose two types of perceptual losses for image deblurring by leveraging the blur-sensitive feature representation of ViT. 

\subsection{Vision Transformer} \label{preliminaries}

Since vision transformers have been widely used as backbone networks in recent studies, we first simply review the overview of ViT. For the 2D image $I$, ViT projects $I$ as a sequence of 1D patch embeddings $T=\{t_i \mid i=1,...,n\}$, where $n$ is the total number of patches. Each patch embedding $t_i$ corresponds to a fixed-size patch of $I$. In addition, position embeddings are added to the patch embeddings to retain positional information about patches. In addition to position embeddings, a patch embedding used for classification is also included as an additional learnable embedding, \ie \verb|[CLS]|, which serves as the global image representation. We refer to complete embeddings as a set of tokens $T=\{t_i \mid i= \verb|CLS|, 1,...,n\}$.

Then, the tokens $T$ are passed through the transformer encoder consisting of a stack of transformer layers. For each transformer layer, there are normalization layers (LN), multi-head self-attention (MSA) modules, and multi-layer perceptron (MLP) blocks. The output of tokens in $(l+1)$-th transformer layer can be formulated as:
\begin{gather}
    T^{l+1} = MSA\left (LN(T^{l}) \right ) + T^{l} \notag \\ 
    T^{l+1} = MLP\left (LN(T^{l+1}) \right ) + T^{l+1}.
\end{gather}
Precisely, in MSA block, tokens are projected into queries, keys, and values:
\begin{gather}
    Q^{l+1} = T^{l} \cdot W_q^{l+1} \notag \\
    K^{l+1} = T^{l} \cdot W_k^{l+1} \notag \\
    V^{l+1} = T^{l} \cdot W_v^{l+1},
\end{gather}
where $W_q^{l+1}, W_k^{l+1}, and W_v^{l+1}$ are learnable weights.

When feeding an image into a pretrained ViT model, we can obtain a set of feature representation $\left \{ T^{l+1}, Q^{l+1}, K^{l+1}, V^{l+1} \right \}$ w.r.t. its token, query, key, and value, respectively. In our framework, we leverage MAE, in which the model can generate feature representation using part of image patches, that is, the number of $T^{l+1}$ can be smaller than $n$.

\subsection{In-depth Transformer Properties}

To verify the effectiveness of transformer properties for image deblurring, we show an example in Fig. \ref{fig:similarity map}, which demonstrates that ViT features can provide fine-grained blur-sensitive guidance at higher spatial resolution. In this figure, we visualize the features using three ViTs: a supervised ViT, a self-supervised DINO, and an MAE. In the realm of computer vision, the introduction of the supervised ViT has marked a significant departure from traditional CNNs, offering a more adaptable and efficient method for processing various image data \cite{dosovitskiy2020vit}. Building on the foundations of ViTs, the development of DINO has propelled self-supervised learning forward by enabling the extraction of rich and meaningful representations from unlabeled data, thereby enhancing the autonomy and adaptability of machine learning models \cite{caron2021emerging}. Additionally, the emergence of MAE has introduced an innovative self-supervised learning technique, which improves the understanding and efficiency of data processing by masking a portion of the input data and tasking the model with reconstructing the obscured information \cite{MaskedAutoencoders2021}. 

It is observed that the similarity heatmaps in a pair of blurred and sharp images are different, indicating that the blur can distort the original global self-similarity. As shown in Fig. \ref{fig:similarity map}(a), the highlighted position marked by $\color{chromeyellow}\star$ in the clear image shows a slight difference compared to the neighborhood regions, while the difference is hard to distinguish in the blurred image. Fortunately, the difference can be significantly revealed by the ViT features, especially the MAE \cite{MaskedAutoencoders2021} shown in Fig. \ref{fig:similarity map}(d). In the similarity heatmap provided by MAE, the features at the marked position show highly similarity to a large area of blurry image. On the contrary, the features of clear image at same marked position show low similarity to neighboring areas.

Based on the observation, we can derive that minimizing the distance of transformer features provides a novel supervision for image deblurring. Note that the MAE features show more salient semantic attention than the other two networks. On this basis, we employ the MAE to preserve the global self-similarity that is used to measure the disparity between the deblurred image and the reference image. Furthermore, we conduct an ablation study to analyze the effectiveness of the three types of ViT features in Sec. \ref{sec: types of vit}.

\subsection{Local MAE Perceptual Loss}\label{sec:local}

% In image deblurring task, most state-of-the-art (SoTA) deep-learning based deblurring methods focus on designing a network architecture that takes the blurry image as input and outputs the recovered image. 
Regarding the objective of optimization, the model minimizes the distance in spatial space between the reconstructed image $I_{recon}$ and the reference image $I_{ref}$:
\begin{equation}
    \mathcal{L}_{deblur} = h(I_{recon}, I_{ref}),
    \label{Eq: basic deblur}
\end{equation}
where $h(\cdot)$ denotes a metric function that used to measure the distance between the restored image and the reference image. The metric $h(\cdot)$ function varies according to the image restoration method employed (see Sec. \ref{Sec: defocus blurring} and Sec. \ref{Sec: motion blurring} for detailed settings). 
To obtain better quality of the reconstruction image, it is critical to improve the perceptual score of recovered images. On this basis, we add a perceptual loss term into Eq. \ref{Eq: basic deblur} using feature representation of transformer:
\begin{equation}
    \mathcal{L}_{total} = \mathcal{L}_{deblur} + \lambda \mathcal{L}_{percep},
    \label{Eq: final loss}
\end{equation}
where $\lambda$ is a balance factor that controls the importance of the regularization term $\mathcal{L}_{percep}$, which includes both $\mathcal{L}_{percep}^{local}$ and $\mathcal{L}_{percep}^{global}$. Details will be provided below.
% our perceptual objective is to measure perceptual quality of recovered images using feature representation of transformer. 
%The key to perceptual loss is how to use the feature representation to measure the perceptual quality of deblurred image. 
% The key to perceptual loss is how to exert effectiveness of feature representation into image deblur task. 
The key to perceptual loss is how to use the feature representation to measure the perceptual quality of the deblurred image. Intuitively, we build a local perceptual loss using MAE feature representation inspired by the VGG-based perceptual loss \cite{johnson2016perceptual}, as shown in Fig. \ref{fig:objectives}(a). We fed the reconstructed image $I_{recon}$ and the reference image $I_{ref}$ into pretrained MAE to extract the representations, as described in Sec. \ref{preliminaries}. Then, we define the local perceptual loss as the Euclidean distance between feature representations:
\begin{equation}
    \mathcal{L}_{percep}^{local} = \left \| \mathcal{F}^{l}(I_{recon}) - \mathcal{F}^{l}(I_{ref}) \right \|_1,
    \label{Eq. local}
\end{equation}
where $\mathcal{F}^l$ denotes the MAE backbone, which produces representations in set $\left \{ T^l, Q^l, K^l, V^l \right \}$. Different from the CNN, the shadow layers of the transformer incline to capture local semantic information, while the deeper layers favor to present the global semantic information. Therefore, it is worth noting that the hyperparameter $l$ plays a key role in the balance of local and global semantic information. The smaller $l$, the less perceptual information is extracted by the MAE. The bigger the $l$, the less attention the loss $\mathcal{L}_{percep}^{local}$ pays to quantitative scores of deblurred images. Please refer to Sec. \ref{sec: encode of mae} for more analysis about the impact of hyperparameter $l$.

\subsection{Global Distribution Perceptual Loss}
\label{sec:dist}

\begin{figure*}[htbp]
    \centering
    \includegraphics[width=18cm]{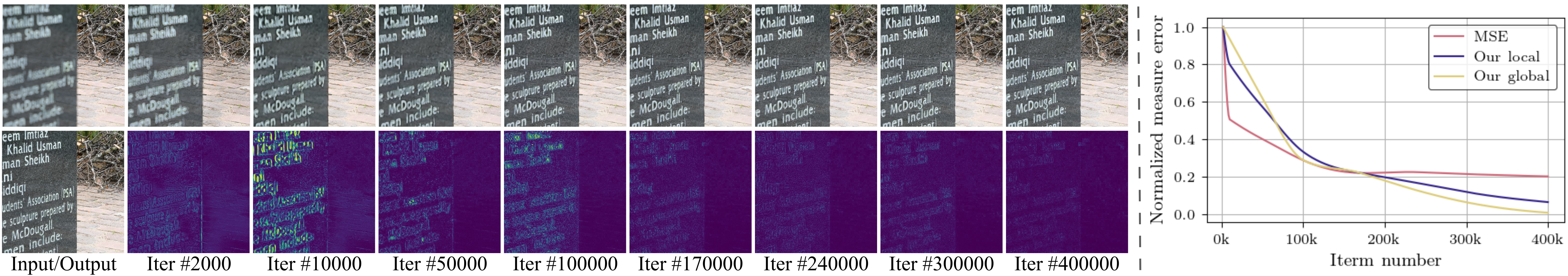}
    \caption{ Illustration of the prediction error measured by quantitative and perceptual measurement at different iterations of deblurring. To highlight the improvement in deblurring results, we show the residual maps that reflect the difference in deblurring results between the current iteration and the previous iteration. }
    \vspace{-10pt}
    \label{fig:error analysis}
    
    % \vspace{-0.5cm}
\end{figure*}

Apart from computing the distance of feature vectors in Euclidean space, we also compare feature distributions between the reconstructed image and reference image based on optimal transport theory. For two sets of feature distributions, the goal of optimal transport is to minimize the cost of transporting the input into the reference distribution. When the cost is defined, the optimal transport establishes a distance between distributions, \ie, p-Wasserstein distance \cite{panaretos2019statistical}. In other words, our proposed distribution perceptual loss is equal to the p-Wasserstein distance between the reconstruction and reference distributions.

Let $\mathcal{F}^l(I_{recon})$ and $\mathcal{F}^l(I_{ref})$ denote the extracted features of the reconstructed image $I_{recon}$ and the reference image $I_{ref}$ in the $l$-th layer of the MAE encoder. To better illustrate the properties of the transformer feature, we take the token features $T^l_{I_{recon}}, T^l_{I_{ref}}$ from $\mathcal{F}^l(I_{recon})$ and $\mathcal{F}^l(I_{ref})$ as the example in the next. The token features $T^l$ consist of $n+1$ token $t_i^l\in \mathbb{R}^{d}$, where each token $t_i^l$ embeds the global topological relations \cite{cao2022understand}. Furthermore, inverting the specific token in the last layer can generate images similar to the input \cite{tumanyan2022splicing}. It demonstrates that each token can be regarded as a set of samples drawn from the distribution.

Based on above considerations, we compare the two distributions with token features, \ie, $\{t_i, k_i, q_i, v_i\}$:
\begin{equation}
    W_p^p(\bm{u}, \bm{v}) = (\sum_{i=1}^d \left \vert u_{i_s} - v_{i_s} \right \vert^p)^{\frac{1}{p}},
    \label{Eq. wasserstein}
\end{equation}
where $\bm{u}=\{u_i\}_{i=1}^{d}$ denotes a set of 1D features, and $u_{i_s}$ represents an ordered sample sorted in ascending order within $\bm{u}$, \ie, $u_{i_s} < u_{i_{s+1}}$ (and similarly for $v_{i_s}$). This implies that, for the comparison of one-dimensional empirical distributions, we can calculate the Wasserstein distance simply by sorting the samples. After computing the Wasserstein distances of all tokens, we can obtain the distribution perceptual loss that is expressed as the sum of Wasserstein distances over the $\mathcal{F}^l(I_{recon})$ and $\mathcal{F}^l(I_{ref})$ :
\begin{equation}
    \mathcal{L}_{percep}^{global} = \sum_{i = {1,...,n, \small \verb|CLS|}} W_p^p(\mathcal{F}_i^l(I_{recon}), \mathcal{F}_i^l(I_{ref})),
    \label{Eq: global loss}
\end{equation}
where $\mathcal{F}_i^l(I_{recon})$ is one of the extracted features from $\mathcal{F}^l(I_{recon})$, \ie, $\{t_i, k_i, q_i, v_i\}$; and similarly for $\mathcal{F}_i^l(I_{ref})$.
In Fig. \ref{fig:objectives}(b), we show an example to explain how to measure the distance between the reconstruction and reference distributions.

As shown in Fig. \ref{fig:error analysis}, we present an example to illustrate the effectiveness of losses in the deblurring process. For traditional $\ell_2$ loss, it is hard to reduce the measurement error as the number of iterations increases. Especially, at the last few iterations (\ie, iteration 240k-400k), the values on the y-axis tend to be consistent. While we found that the measured error continues to decrease under the two proposed types of perceptual losses measurement in Sec \ref{sec:local} and Sec \ref{sec:dist}. It indicates that the perceptual quality of deblurred results continues to improve, even though the quantitative loss ($\ell_2$) suffers from premature convergence. Furthermore, we show the residual maps to highlight the differences among multiple iterations. The residual maps in the last few iterations depict that characters containing more semantic information tend to receive more attention.

\section{Experiments}\label{sec4}

In this section, we introduce our experiments settings and implementation details before evaluating the performance of proposed perceptual losses in various image deblurring scenarios, including defocus deblurring and motion deblurring. Then, we conduct ablation studies to further demonstrate the benefit of perceptual loss with different transformer properties. 

\subsection{Settings and Details} 

\subsubsection{Dataset} 
\label{Sec. dataset}
We evaluate the performance of our method on three representative datasets for defocus deblurring and motion blurring, respectively. The datasets are as follows:

\noindent \textbf{DPDD.} The DPDD dataset \cite{abuolaim2020defocus} provides real defocus blur and all-in-focus pairs, captured with a DSLR camera. There are 500 indoor/outdoor scenes in the DPDD dataset where each scene contains two blurred sub-aperture views (left and right views) stored in lossless 16-bit depth. The DPDD dataset is divided into three sub-datasets: training, validation, and testing datasets, consisting of 350, 74, and 76 pairs, respectively. Note that we can easily obtain the single defocus blur image by averaging the two blur sub-aperture views.

\noindent \textbf{GoPro.} The GoPro dataset \cite{nah2017deep} is a synthetic dataset in which the synthetic blurry images are simulated from high frame rate videos captured by a high-speed camera, \ie, GoPro. By averaging consecutive short-exposure frames, blurry images that simulate blurs caused by slow shutter speed contain more realistic complex dynamic motion. The GoPro dataset contains 3214 blurry/sharp image pairs at 1280$\times$720 resolution, of which 2103 are used for training and the remaining are reserved for evaluation. 

\noindent \textbf{HIDE.} The HIDE dataset \cite{shen2019human} is also a synthetic dataset that is specifically designed for human-aware motion deblurring. The images are collected from a wide range of daily scenes, covering diverse human motions and complex backgrounds. In this paper, we employ the test dataset containing 2025 pairs of images to evaluate in the motion blur task.

% \noindent \textbf{RealBlur.} The RealBlur dataset \cite{rim2020real} is collected in real-world conditions. The blurry and sharp images are simultaneously captured by an specific image acquisition system that addresses the geometric alignment challenge well. The RealBlur dataset consists of two subsets, \ie, RealBlur-R and RealBlur-J. RealBlur-R dataset is generated from camera raw images and then postprocessed by the white balance, demosaicing, and denoising. RealBlur-J dataset originates from the JPEG formats in the camera. Each dataset contains 4,738 pairs of images over 232 different scenes. 

% For the GoPro dataset, we use it as-is without additional preprocessing.

\subsubsection{Implementation Details}

Before training the image defocus deblurring, we first preprocess the DPDD dataset as suggested by DPDNet \cite{abuolaim2020defocus}. Specifically, we crop the 1680$\times$1120 image into 512$\times$512 patches by sliding a window striding with 60\% overlap. Then, we discard about 30\% patches as they are too homogeneous, resulting in adversely affect the training \cite{park2017unified}. For the GoPro dataset, we use it as is without extra preprocess.

\begin{figure*}[htbp]
    \centering
    \includegraphics[width=18.1cm]{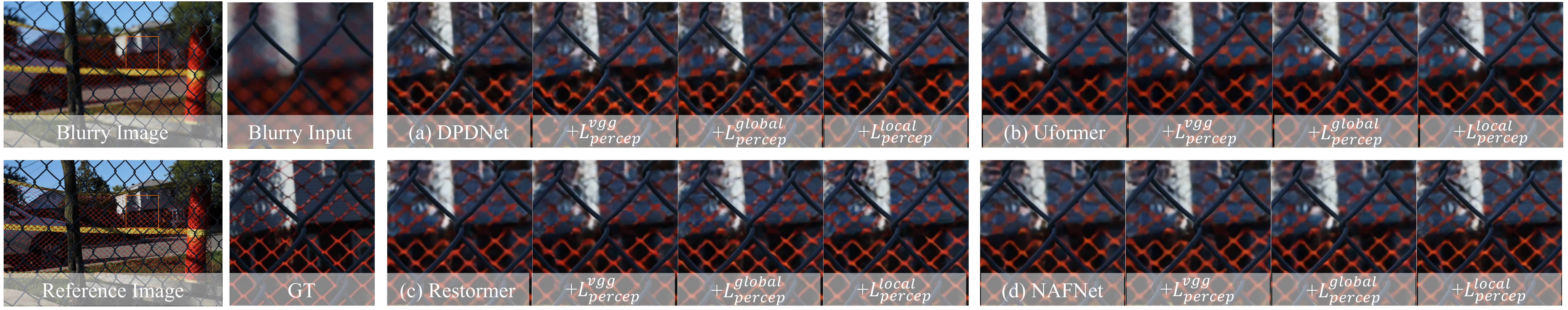}
    \caption{ A visual example on the dual-pixel DPDD dataset. For convenience, we show only the left view as the blurry image. From (a) to (d), we present cropped highlighted deblurring results to validate the effectiveness of proposed perceptual losses. }
    \label{fig:dualpixel deblurring}
    \vspace{-3mm}
\end{figure*}

\begin{figure*}[htbp]
    \centering
    \includegraphics[width=18cm]{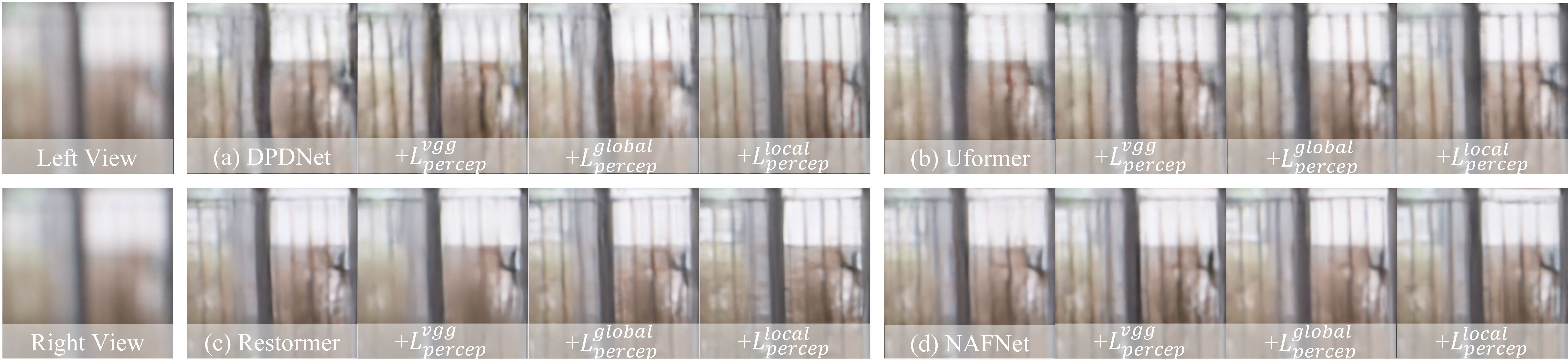}
    \caption{ Visual comparison of dual-pixel blurry images on the DLDP test dataset. Compared to the baseline, the models equipped with our proposed loss generate sharper and visually-faithful result. }
    \vspace{-10pt}
    \label{fig: dldp}
\end{figure*}

In training, we empirically use token features of the MAE encoder as transformer features, and set the layer parameter $l$ to 5 by default. For the balance factor $\lambda$, we set $\lambda=1$ in Eq. \ref{Eq: final loss} and $\lambda=1e-5$ in Eq. \ref{Eq: global loss}. In particular, we set the mask ratio of MAE to 50\% in the case of the local MAE perceptual loss. For a fair comparison, we train and test the method with the same training dataset under the Pytorch framework. To this end, we reimplement some methods under the Pytorch framework. In this process, we crop the image into patches of (256 $\times$ 256) and initiate the learning rate at $2e-4$. The rate is halved after every 40 epochs, progressively diminishing to a final rate of $2e-5$. The procedure is conducted over a total of 200 epochs. For optimization, the Adam algorithm is utilized with a consistent weight decay factor set at $5e-4$ across all experimental runs. The standard batch size is set to 4 for all methods with the exception of the Restormer, for which the batch size is reduced to 2.  As depicted in Eq. \ref{Eq: global loss}, the global perceptual loss $\mathcal{L}_{{percep}}^{{global}}$ incorporates a hyperparameter p, which is assigned a value of 2. Considering the RTX 3090 GPU used with 24G memory capacity, we tune down the batch size of some approaches to fit the maximum memory limitation and avoid running out of memory.

% The two sub-aperture blurred views are concatenated into a 6-channel image as input.

\newcommand{\halftablespace}{\,}
\newcommand{\tbold}[1]{\pmb{#1}}
\newcommand{\tunder}[1]{\underline{#1}}

\newcommand{\vershiftY}{0.25}
\newcommand{\vershiftYtwo}{0.4}

\begin{table*}[htbp]
\footnotesize
% \footnotesize
\caption{ Results for the defocus deblurring task on the dual-pixel DPDD dataset \cite{abuolaim2020defocus}. The \tbold{bold} numbers indicate the best results and the second bests are marked by \tunder{underlines}.}
% \vspace{-5pt}
\label{tab:dualpixel deblurring}
\begin{center}
    \renewcommand{\arraystretch}{1.0}
    \setlength\tabcolsep{2.25pt}{
    \begin{tabular}{@{\,\,}>{\raggedright}p{0.074\textwidth}|>{\raggedright}p{0.073\textwidth}|p{0.040\textwidth}p{0.043\textwidth}p{0.043\textwidth}p{0.043\textwidth}p{0.043\textwidth}|p{0.040\textwidth}p{0.043\textwidth}p{0.043\textwidth}p{0.043\textwidth}p{0.043\textwidth}|p{0.040\textwidth}p{0.043\textwidth}p{0.043\textwidth}p{0.043\textwidth}p{0.043\textwidth}@{\,\,}}
    \multirow{2}{*}{Method} & \multirow{2}{*}{\,\,\,\,\,\,\,Loss} &  \multicolumn{5}{c|}{\emph{Indoor}} & \multicolumn{5}{c|}{\emph{Outdoor}} & \multicolumn{5}{c}{\emph{Combined}}\\
     & &PSNR & SSIM & MAE & LPIPS & NIQE &PSNR & SSIM & MAE & LPIPS & NIQE  &PSNR & SSIM & MAE & LPIPS & NIQE \\
    \hline
    DPDNet    & \,\,\,\,\,\,\,\,\,\, -                     & 28.05 & \tunder{0.8644} & 0.0262 & 0.1341 & 5.078 & 22.98 & \tunder{0.7364} & \tunder{0.0514} & 0.1711 & 4.370 & 25.45 & \tunder{0.7987} & \tunder{0.0391}  & 0.1531 & 4.715 \\
    & +$\mathcal{L}_{percep}^{vgg}$      & 28.06 & 0.8615 & 0.0264 & 0.0920 & \tbold{4.718} & 22.85 & 0.7355 & 0.0522 & 0.1376 & \tbold{4.046} & 25.39 &0.7969 & 0.0396 & 0.1154 & \tbold{4.373} \\
    & +$\mathcal{L}_{percep}^{local}$      & $\pmb{28.57}$ & \tbold{0.8652} & \tbold{0.0258} & \tbold{0.0766} & \tunder{4.990} & \tbold{23.35} & \tbold{0.7375} & \tbold{0.0501} & \tbold{0.1114} & 4.338 & \tbold{25.88} & \tbold{0.7997} & \tbold{0.0382} & \tbold{0.0945} & 4.656 \\
    & +$\mathcal{L}_{percep}^{global}$         & \tunder{28.11} & 0.8593 & \tunder{0.0261} & \tunder{0.0902} & 4.751 & \tunder{23.02} & 0.7343 & \tunder{0.0514} & \tunder{0.1358} & \tunder{4.129} & \tunder{25.50}   & 0.7952   & \tunder{0.0391}  & \tunder{0.1049}  & \tunder{4.432} \\
    \arrayrulecolor{gray}
    \hline
    \arrayrulecolor{black}
    Uformer    & \,\,\,\,\,\,\,\,\,\, -                   & 28.59 & 0.8733 & 0.0251 & 0.1279 & 5.123 & 23.48 & 0.7524 & 0.0493 & 0.1672 & 4.306 & 25.97 & 0.8111  & 0.0375  & 0.1480  & 4.704 \\
    & +$\mathcal{L}_{percep}^{vgg}$      & 28.61 & 0.8760 & 0.0253 & 0.1085 & 5.042 & 23.56 & 0.7544 & 0.0490 & 0.1519 & 4.190 & 26.02 & 0.8135 & 0.0375 & 0.1308 & 4.605 \\
    & +$\mathcal{L}_{percep}^{local}$      & \tbold{28.84} & \tbold{0.8788} & \tbold{0.0246} & \tbold{0.0771} & \tunder{4.865} & \tbold{23.68} & \tbold{0.7611} & \tbold{0.0482} & \tbold{0.1187} & \tunder{4.083} & \tbold{26.20} & \tbold{0.8184} & \tbold{0.0367} & \tbold{0.0985}  & \tunder{4.464} \\
    & +$\mathcal{L}_{percep}^{global}$         & \tunder{28.73} & \tunder{0.8763} & \tunder{0.0248} & \tunder{0.0820} & \tbold{4.685} & \tunder{23.56} & \tunder{0.7551} & \tunder{0.0489} & \tunder{0.1265} & \tbold{3.950} & \tunder{26.08} & \tunder{0.8141} & \tunder{0.0371} & \tunder{0.1048} & \tbold{4.307} \\
    \arrayrulecolor{gray}
    \hline
    \arrayrulecolor{black}
    Restormer   & \,\,\,\,\,\,\,\,\,\, -                   & 29.06 & \tunder{0.8850} & 0.0233 & 0.1318 & 5.451 & 23.67 & 0.7647 & 0.0481 & 0.1506 & 4.668 & 26.29  & \tunder{0.8232} & 0.0360 & 0.1414 & 5.049 \\
    & +$\mathcal{L}_{percep}^{vgg}$      & \tbold{29.17} & \tbold{0.8858} & \tbold{0.0231} & 0.1042 & 5.203 & 23.66 & 0.7628 & 0.0482 & 0.1323 & 4.410 & 26.34   & 0.8227  & \tunder{0.0359}  & 0.1186  & 4.796 \\
    & +$\mathcal{L}_{percep}^{local}$      &\tbold{29.17} & 0.8839 & 0.0235 & \tbold{0.0689} & \tbold{4.929} & \tbold{23.84} & \tbold{0.7660} & \tbold{0.0473} & \tbold{0.0999} & \tunder{4.248} & \tbold{26.43} & \tbold{0.8234}  & \tbold{0.0357}  & \tbold{0.0848}  & \tunder{4.579} \\
    & +$\mathcal{L}_{percep}^{global}$         &\tunder{29.11} & 0.8829 & \tunder{0.0232} & \tunder{0.0775} & \tunder{4.934} & \tunder{23.72} & \tunder{0.7651} & \tunder{0.0480} & \tunder{0.1113} & \tbold{4.231} & \tunder{26.35} & 0.8224 & \tunder{0.0359}  & \tunder{0.0949}  &  \tbold{4.573} \\ 
    \arrayrulecolor{gray}
    \hline
    \arrayrulecolor{black}
    NAFNet       & \,\,\,\,\,\,\,\,\,\, -                  & \tunder{29.21} & \tbold{0.8875} & \tbold{0.0234} & 0.1230 & 5.295 & 23.81 & 0.7665 & 0.0480 & 0.1558 & 4.491 & 26.44 & \tunder{0.8254} & 0.0360 & 0.1399 & 4.883 \\
    & +$\mathcal{L}_{percep}^{vgg}$      & 29.09 & \tunder{0.8857} & 0.0238 & 0.1122 & 5.239 & 23.81 & \tunder{0.7678} & 0.0480 & 0.1453 & 4.371 & 26.38 & 0.8251 & 0.0362 & 0.1292 & 4.793 \\
    & +$\mathcal{L}_{percep}^{local}$      &\tbold{29.24} & 0.8848 & \tunder{0.0235} & \tbold{0.0672} & \tbold{4.832} & \tbold{23.95} & \tbold{0.7701} & \tbold{0.0468} & \tbold{0.1024} & \tbold{4.091} & \tbold{26.53} & \tbold{0.8259} & \tbold{0.0355} & \tbold{0.0852} & \tbold{4.451} \\
    & +$\mathcal{L}_{percep}^{global}$         & 29.17 & 0.8850 & \tunder{0.0235} & \tunder{0.0764} & \tunder{4.910} & \tunder{23.86} & 0.7671 & \tunder{0.0477} & \tunder{0.1168} & \tunder{4.253} & \tunder{26.45} & 0.8245 & \tunder{0.0359} & \tunder{0.0971} & \tunder{4.572} \\ 
    \end{tabular}%
    }
\end{center}
\end{table*}

\begin{table*}[htbp]
% \footnotesize
\footnotesize
\caption{ Results for defocus deblurring task on the DPDD dataset \cite{abuolaim2020defocus}. The \tbold{bold} numbers indicate the best results and the second bests are marked by \tunder{underlines}.
% Each blurred input is a 3-channel blurred image which contains more blur amount than its sub-aperture blur view.
}
\vspace{-5pt}
\label{tab:single-deblurring}
\begin{center}
    \renewcommand{\arraystretch}{1.0}
    \setlength\tabcolsep{3.25pt}{
    \begin{tabular}{@{\,\,}>{\raggedright}p{0.074\textwidth}|>{\raggedright}p{0.073\textwidth}|p{0.040\textwidth}p{0.043\textwidth}p{0.043\textwidth}p{0.043\textwidth}p{0.043\textwidth}|p{0.040\textwidth}p{0.043\textwidth}p{0.043\textwidth}p{0.043\textwidth}p{0.043\textwidth}|p{0.040\textwidth}p{0.043\textwidth}p{0.043\textwidth}p{0.043\textwidth}p{0.043\textwidth}@{\,\,}}
    \multirow{2}{*}{Method} & \multirow{2}{*}{\,\,\,\,\,\,\,Loss} &  \multicolumn{5}{c|}{\emph{Indoor}} & \multicolumn{5}{c|}{\emph{Outdoor}} & \multicolumn{5}{c}{\emph{Combined}}\\
     & &PSNR & SSIM & MAE & LPIPS & NIQE &PSNR & SSIM & MAE & LPIPS & NIQE  &PSNR & SSIM & MAE & LPIPS & NIQE \\
    \hline
    
    DPDNet      & \,\,\,\,\,\,\,\,\,\, -              & 27.46 & \tunder{0.8420} & 0.0280 & 0.1675 & 5.3151 & 22.26 & 0.6920 & 0.0553 & 0.2167 & 4.6339 & 24.79 & 0.7650 & 0.0420 & 0.1928 & 4.9655 \\ 
    & +$\mathcal{L}_{percep}^{vgg}$          &27.39 & \tbold{0.8426} & 0.0283 & 0.1138 & \tunder{4.8428} & 22.21 & 0.6920 & 0.0555 & 0.1801 & \tunder{4.1922} & 24.73 & 0.7653 & 0.0423 & 0.1478 & \tunder{4.5089} \\
    & +$\mathcal{L}_{percep}^{local}$      &\tunder{27.53} & 0.8418 & \tbold{0.0277} & \tunder{0.1071} & 4.8961 & \tbold{22.39} & \tbold{0.6954} & \tbold{0.0544} & \tunder{0.1716} & 4.2747 & \tbold{24.89} & \tbold{0.7667} & \tbold{0.0414} & \tunder{0.1402} & 4.5772 \\
    & +$\mathcal{L}_{percep}^{global}$         &\tbold{27.55} & 0.8398 & \tunder{0.0278} & \tbold{0.1070} & \tbold{4.6802} & \tunder{22.29} & \tunder{0.6949} & \tunder{0.0551} & \tbold{0.1708} & \tbold{4.0564} & \tunder{24.85} & \tunder{0.7654} & \tunder{0.0418} & \tbold{0.1398} & \tbold{4.3601}  \\
    \arrayrulecolor{gray}
    \hline
    \arrayrulecolor{black}
    
    Uformer    & \,\,\,\,\,\,\,\,\,\, -                 & 27.90 & 0.8573 & 0.0265 & 0.1427 & 5.1928 & 22.56 & 0.7187 & 0.0538 & 0.1873 & 4.3566 & 25.16 & 0.7862 & 0.0405 & 0.1656 & 4.7637 \\
    & +$\mathcal{L}_{percep}^{vgg}$          & 27.86 & \tunder{0.8604} & 0.0265 & 0.1298 & 5.1113 & \tunder{22.68} & \tunder{0.7213} & \tunder{0.0530} & 0.1787 & 4.2743 & 25.20 & \tunder{0.7890} & 0.0401 & 0.1549 & 4.6818 \\
    & +$\mathcal{L}_{percep}^{local}$      & \tbold{28.17} & \tbold{0.8634} & \tbold{0.0259} & \tbold{0.0797} & \tunder{4.8137} & \tbold{22.85} & \tbold{0.7257} & \tbold{0.0519} & \tbold{0.1297} & \tbold{4.0440} & \tbold{25.44} & \tbold{0.7928} & \tbold{0.0392} & \tbold{0.1054} & \tbold{4.4187} \\
    & +$\mathcal{L}_{percep}^{global}$         & \tunder{27.94} & 0.8568 & \tunder{0.0264} & \tunder{0.0944} & \tbold{4.8131} & 22.67 & 0.7184 & \tunder{0.0530} & \tunder{0.1526} & \tunder{4.1799} & \tunder{25.24} & 0.7857 & \tunder{0.0400} & \tunder{0.1243} & \tunder{4.4882}  \\
    \arrayrulecolor{gray}
    \hline
    \arrayrulecolor{black}
    Restormer   & \,\,\,\,\,\,\,\,\,\, -                & 28.27 & 0.8631 & \tunder{0.0254} & 0.1435 & 5.4370 & 22.91 & 0.7143 & \tunder{0.0511} & 0.2032 & 4.9121 & 25.52 & 0.7868 & \tbold{0.0386} & 0.1741 & 5.1676 \\
    & +$\mathcal{L}_{percep}^{vgg}$          &\tunder{28.32} & 0.8626 & 0.0256 & 0.1411 & 5.3994 & 22.89 & 0.7143 & \tbold{0.0510} & 0.1908 & 4.8744 &   25.54 & 0.7865 & \tbold{0.0386} & 0.1666 & 5.1300 \\
    & +$\mathcal{L}_{percep}^{local}$      &\tbold{28.56} & \tbold{0.8688} & \tbold{0.0253} & \tbold{0.0854} & \tbold{4.9113} & \tunder{22.94} & \tunder{0.7220} & 0.0514 & \tbold{0.1307} & \tbold{4.4031} & \tbold{25.68} & \tbold{0.7935} & \tunder{0.0387} & \tbold{0.1086} & \tbold{4.6505} \\
    & +$\mathcal{L}_{percep}^{global}$         &\tunder{28.32} & \tunder{0.8645} & 0.0256 & \tunder{0.1102} & \tunder{5.1762} & \tbold{23.00} & \tbold{0.7233} & \tunder{0.0511} & \tunder{0.1646} & \tunder{4.6384} & \tunder{25.59} & \tunder{0.7920} & \tunder{0.0387} & \tunder{0.1381} & \tunder{4.9002}  \\ 
    \arrayrulecolor{gray}
    \hline
    \arrayrulecolor{black}
    
    NAFNet      & \,\,\,\,\,\,\,\,\,\, -                & 28.16 & 0.8651 & \tunder{0.0257} & 0.1365 & 5.3256 & 22.75 & \tunder{0.7231} & 0.0526 & 0.1861 & 4.5713 & 25.38 & 0.7922 & 0.0395 & 0.1619 & 4.9385 \\
    & +$\mathcal{L}_{percep}^{vgg}$          & \tunder{28.20} & \tbold{0.8671} & \tunder{0.0257} & 0.1250 & 5.2839 & 22.73 & 0.7230 & 0.0528 & 0.1751 & 4.3924 & 25.39 & \tbold{0.7932} & 0.0396 & 0.1507 & 4.8265 \\
    & +$\mathcal{L}_{percep}^{local}$      & \tbold{28.42} & \tunder{0.8664} & \tbold{0.0249} & \tbold{0.0781} & \tbold{4.9439} & \tbold{22.88} & \tbold{0.7237} & \tbold{0.0517} & \tbold{0.1299} & \tbold{4.0781} & \tbold{25.58} & \tunder{0.7931} & \tbold{0.0387} & \tbold{0.1047} & \tbold{4.4996} \\
    & +$\mathcal{L}_{percep}^{global}$         & 28.18 & 0.8612 & \tunder{0.0257} & \tunder{0.0890} & \tunder{4.9993} & \tunder{22.80} & 0.7198 & \tunder{0.0525} & \tunder{0.1468} & \tunder{4.3917} & \tunder{25.42} & 0.7887 & \tunder{0.0395} & \tunder{0.1187} & \tunder{4.6875} \\ 
    \end{tabular}%
    }
    \vspace{-10pt}
\end{center}
\end{table*}

% \renewcommand{\thetable}{R\arabic{table}}
% \begin{table}[htbp]
% \caption{Results on the Rain100H dataset \cite{yang2017deep} for single image derain task. The \tbold{bold} numbers indicate the best results and the second bests are marked by \tunder{underlines}.}
% \label{tab:image rain}
% \normalsize
% \begin{center}
%     \renewcommand{\arraystretch}{1.2}
%     \setlength\tabcolsep{10.5pt}{
%     \begin{tabular}{l|l|c|c}
%     Method & Loss & PSNR & SSIM \\
%     \hline
%     Uformer & baseline & 30.92 & 0.9028 \\
%             & +$\mathcal{L}_{percep}^{local}$ & \tbold{31.30} & \tbold{0.9093} \\
%             & +$\mathcal{L}_{percep}^{global}$  & \tunder{30.96} & \tunder{0.9030} \\
%     \arrayrulecolor{gray}
%     \hline
%     \arrayrulecolor{black}
%     DRSformer & baseline & 31.25 & 0.9101 \\
%               & +$\mathcal{L}_{percep}^{local}$ & \tunder{31.27} & \tbold{0.9145} \\
%               & +$\mathcal{L}_{percep}^{global}$ & \tbold{31.29} & \tunder{0.9106} \\
%     \arrayrulecolor{gray}
%     \hline
%     \arrayrulecolor{black}    
%     Restormer & normal & 31.54 & 0.9165 \\
%               & +$\mathcal{L}_{percep}^{local}$ & \tunder{31.70} & \tbold{0.9185} \\
%               & +$\mathcal{L}_{percep}^{global}$ & \tbold{31.87} & \tunder{0.9175} \\
%     \end{tabular}
%     }
% \end{center}
% \end{table}

\subsubsection{Evaluation Metrics}
\label{Sec: metric introduce}

In order to comprehensively evaluate the effectiveness of our perceptual losses, we calculate both the traditional quantitative quality metrics and the perceptual metrics. For traditional quantitative quality metrics, we calculate PSNR (higher is better), SSIM (higher is better), and MAE (lower is better) metrics between the recovered image and the reference ground truth. For perceptual metrics, we introduce a non-reference metric NIQE (lower is better) \cite{mittal2012making} and a reference-based metric LPIPS (lower is better) \cite{zhang2018unreasonable}. The NIQE metric is widely used for conditional image generation, indicating the visual realism of generated images. The LPIPS metric is correlated with human perceptual similarity judgments by using deep features in a pretrained VGG.

\subsection{Defocus Deblurring}
\label{Sec: defocus blurring}
\begin{figure*}[htbp]
    \centering
    \includegraphics[width=18.1cm]{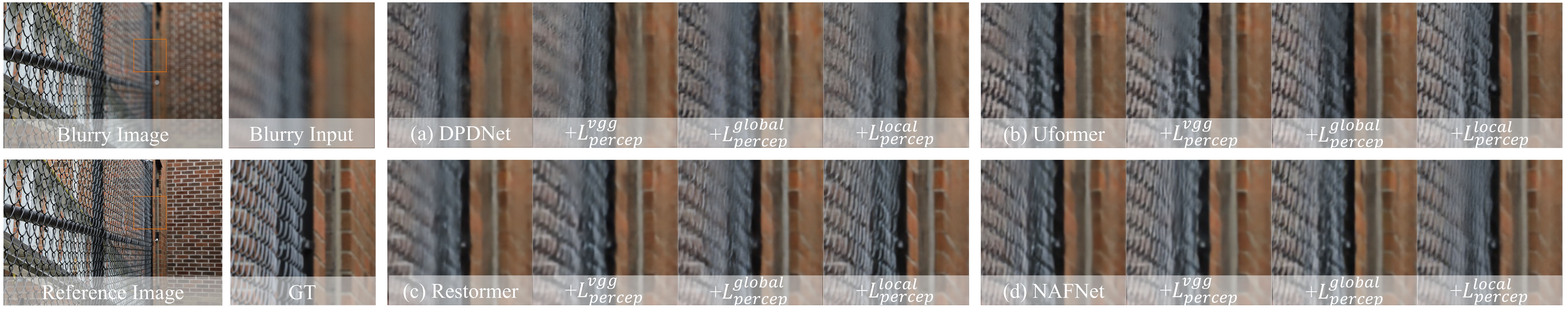}
    \caption{Comparison results of a single blurry image on the DPDD test dataset. From (a) to (d), we present cropped highlighted deblurring results to validate the effectiveness of proposed perceptual losses.}
    \vspace{-10pt}
    \label{fig:single deblurring}
\end{figure*}

We select four learning-based methods as our baselines, including DPDNet \cite{abuolaim2020defocus}, Uformer-T (abbreviated Uformer in the following sections) \cite{wang2022uformer}, Restormer \cite{zamir2022restormer}, and NAFNet \cite{chen2022simple}, which are the most representative and competitive in the literature. For DPDNet and Restormer, we utilize the $\ell_l$ as the metric function $h(\cdot)$ defined in Eq. \ref{Eq: basic deblur}. Similarly, we apply the Charbonnier loss to Uformer and the PSNR loss to NAFNet. We evaluate different network structures with two different cases: (i) two blurry views of the dual-pixel image pair and (ii) single blurry image. For the dual-pixel defocus deblurring, the quantitative evaluation results and objective comparisons are presented in Fig. \ref{fig:dualpixel deblurring}, Fig. \ref{fig: dldp} and Tab. \ref{tab:dualpixel deblurring}. For the single view deblurring, we show visual results in Fig. \ref{fig:single deblurring} and objective results in Tab. \ref{tab:single-deblurring}.

The results in Tabs. \ref{tab:dualpixel deblurring} and \ref{tab:single-deblurring} show that training with local MAE perceptual loss provides the most compelling deblurring results in terms of quantitative scores and perceptual metrics. 
% Replacing computing features in Euclidean space with comparing the feature distributions, 
Replacing the feature representation extracted from VGG with those extracted from MAE, the proposed global loss $\mathcal{L}_{percep}^{global}$ produces competitive perceptual and quantitative results against the classical perceptual loss. Particularly, in term of PSNR, the proposed $\mathcal{L}_{percep}^{local}$ and $\mathcal{L}_{percep}^{global}$ produce significantly better values than the baseline, while showing the superior perceptual performance. On the contrary, model training with classical perceptual loss tends to decrease PSNR scores. Overall, all perceptual losses effectively improve the perceptual quality of deblurred images, where our proposed losses can help the deblurring model recover a more realistic image with better perceptual scores. It demonstrates that in-depth transformer properties are more beneficial for reconstructing high-quality images than the pretrained VGG. In most cases, perceptual losses based on transformer properties lead to performance gains on both quantitation and perception measurement over the baseline.

To further illustrate the advantage of proposed perceptual losses, we show the visual comparison in Figs. \ref{fig:dualpixel deblurring} and \ref{fig:single deblurring}. The results recovered by the model trained with $\mathcal{L}_{percep}^{vgg}$ exhibit obvious artifacts as shown in Fig. \ref{fig:dualpixel deblurring}, \eg, the crossed red lines contain obvious noise perturbation. When fed single blurry input, the model guided by $\mathcal{L}_{percep}^{vgg}$ yields over-smooth deblurring results. On the contrary, the models trained with our proposed losses contains more proper structures with fewer artifacts in Fig. \ref{fig:dualpixel deblurring}. When evaluating on the single image deblurring task, the models recover visibly clearer images and much sharper structures under the supervision of our perceptual losses as shown in Fig. \ref{fig:single deblurring}. As described in Sec. \ref{Sec: metric introduce} the perceptual metric (LPIPS) appears to suggest a better agreement with humans, where the lower LPIPS score refers to better image quality. Overall, we find that our visual performance is consistent with its objective results, where the model equipped with our perceptual losses yields richer details and achieves better performance in perceptual measurement. 

Additionally, we present a visual result on the DLDP dataset to verify the generalization of the proposed perceptual loss, as illustrated in Fig. \ref{fig: dldp}. It is observable that our model, when equipped with the proposed loss functions, tends to yield clearer restored images. In the zoomed-in areas of the figure, it is evident that the images processed with both local and global perceptual losses display sharper edges and richer textures compared to those processed with the VGG perceptual loss models and baselines. It demonstrates that the models trained with the proposed perceptual loss are robust to various scenes.

% Overall, the visual performance advantage is consistent with its objective results.

\subsection{Motion Deblurring}
\label{Sec: motion blurring}
\begin{figure}[htbp]
    \centering
    \includegraphics[width=8.8cm]{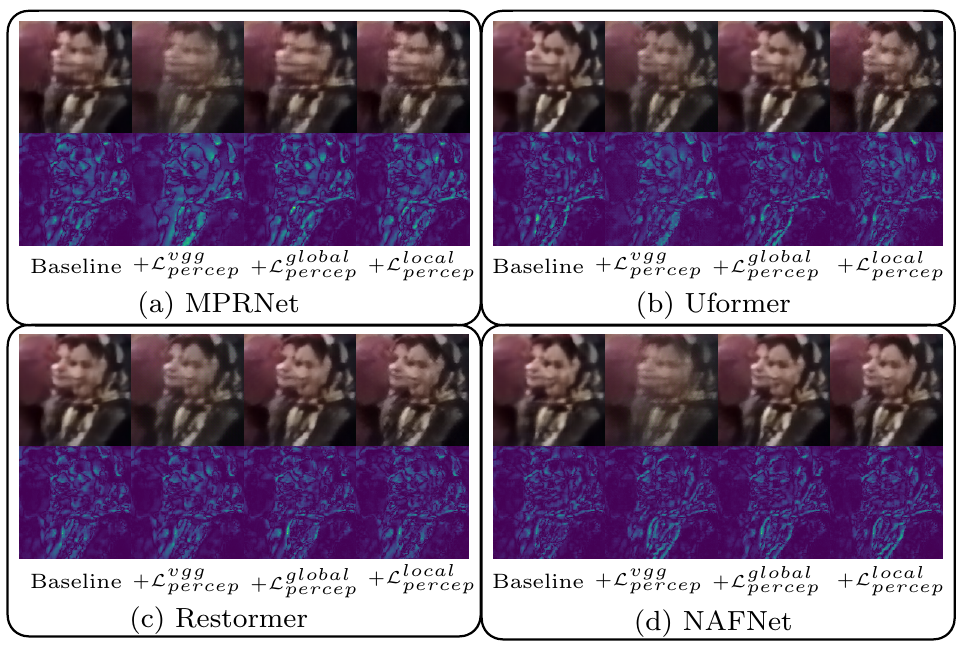}
    \caption{ Visual comparisons of the motion deblurring task on the GoPro test dataset \cite{nah2017deep}. To better highlight the differences, we visualize the residual maps between recovered patches and reference in (a)-(d). }
    \vspace{-10pt}
    \label{fig:motion deblurring}
\end{figure}

To explore the effectiveness of proposed perceptual losses, we evaluate the models on the GoPro and HIDE datasets. In particular, we train the model on the GoPro train dataset and directly test it on HIDE and GoPro test datasets following the experimental settings used in MPRNet \cite{zamir2021multi}. Four representative models are used to evaluate the proposed perceptual losses, including MPRNet \cite{zamir2021multi}, Uformer \cite{wang2022uformer}, Restormer \cite{zamir2022restormer}, and NAFNet \cite{chen2022simple}. In the original four baseline models, they adopt different metric loss functions $h(\cdot)$ : MPRNet is trained with $\ell_1$ and an edge regularization term, Uformer is trained with Charbonnier loss \cite{zamir2020learning}, Restormer is equipped with $\ell_1$, while NAFNet is equipped with PSNR loss \cite{chen2021hinet}. For fair comparison, we add the proposed perceptual loss as a new regularization term and keep other losses the same as the settings of original papers. In Tab. \ref{tab:motion deblurring} and Fig. \ref{fig:motion deblurring}, we present objective comparisons and visual results, respectively.

In objective comparisons, the proposed perceptual losses exhibit advantages over the classical perceptual loss. Specifically, training with the classical perceptual loss $\mathcal{L}_{percep}^{vgg}$, the models (\eg, Uformer, and Restormer) tend to produce the disputing deblurring results where the two perceptual metrics (\ie, LPIPS and NIQE) give inconsistent quality assessment. In addition to perception measurements, the deblurred results generated by the model with classical perceptual loss show worse quantitative scores compared to baseline models w.r.t. MPRNet, Uformer, and Restormer. In contrast, models equipped with our proposed perceptual losses consistently perform better perceptual quality scores than baselines. Furthermore, in term of PSNR, our perceptual losses show competitive performance against the baseline on the MPRNet and go beyond other counterparts on Uformer, Restormer, and NAFNet. Overall, our proposed losses are effective on the motion deblurring task, which is thoroughly demonstrated on the HIDE and GoPro datasets. 

For visual comparisons, we present an example from GoPro test dataset as shown in Fig. \ref{fig:motion deblurring}. The models regularized with $\mathcal{L}_{percep}^{vgg}$ recover low-contrast images, showing a color shift to some extent. To better highlight the missing details, we also provide the visualization of residual maps between deblurring results and ground truth. It can be seen that our perceptual losses, especially for $\mathcal{L}_{percep}^{local}$, have a tendency to yield a perceptually pleasing and sharp images with fewer artifacts. Since both objective and visual comparisons reveal that the proposed perceptual losses are effective for the motion deblurring task, it demonstrates that our losses are robust to various and complex blurred scenes.

\begin{table}[htbp]
\scriptsize
\caption{Results on the GoPro dataset \cite{nah2017deep} and the HIDE \cite{shen2019human} for motion deblurring task. The \tbold{bold} numbers indicate the best results and the second bests are marked by \tunder{underlines}.}
\label{tab:motion deblurring}
\begin{center}
    \renewcommand{\arraystretch}{1.0}
    \setlength\tabcolsep{1.0pt}{
    \begin{tabular}{@{\,\,}>{\raggedright}p{0.056\textwidth}|>{\raggedright}p{0.064\textwidth}|>{\raggedright}p{0.036\textwidth}p{0.043\textwidth}p{0.043\textwidth}p{0.043\textwidth}|p{0.036\textwidth}p{0.043\textwidth}p{0.043\textwidth}p{0.043\textwidth}@{\,\,}}
    \multirow{2}{*}{Method} & \multirow{2}{*}{\,\,\,\,\,\,\,Loss} &  \multicolumn{4}{c|}{\emph{GoPro}} & \multicolumn{4}{c}{\emph{HIDE}}\\
    &  &PSNR & SSIM & LPIPS & NIQE &PSNR & SSIM & LPIPS & NIQE \\
    \hline
    MPRNet         & \,\,\,\,\,\,\,\,\,\, -                & \tbold{28.55}   & \tbold{0.9111}  & 0.1343  & 5.4627  & \tbold{27.25}  & \tbold{0.8847}  & 0.1425 & 4.9244 \\ 
    &  +$\mathcal{L}_{percep}^{vgg}$          & 28.15   & 0.9040  & \tbold{0.0876}  & \tbold{4.7615}  & 27.09  & 0.8792  & \tbold{0.1073}  & \tbold{4.1345}  \\
    &  +$\mathcal{L}_{percep}^{local}$      & 28.47   & \tunder{0.9093}  & \tunder{0.1102}  & \tunder{4.9845}  & 27.12  & 0.8818  & \tunder{0.1250}  & \tunder{4.3741}  \\
    &  +$\mathcal{L}_{percep}^{global}$         & \tunder{28.48}   & \tunder{0.9093}  & 0.1145  & 5.1228  & \tunder{27.18}  & \tunder{0.8824}  & 0.1282 & 4.6510  \\
    \arrayrulecolor{gray}
    \hline
    \arrayrulecolor{black}
    Uformer        & \,\,\,\,\,\,\,\,\,\, -                & 30.21   & \tunder{0.9343}  & 0.1135  & 5.5583  & 28.48  & \tunder{0.9079} & 0.1296  & 4.9227 \\
    &  +$\mathcal{L}_{percep}^{vgg}$          & 29.72   & 0.9197  & \tunder{0.0694}  & 5.9382  & 28.09 & 0.8820 & \tunder{0.0920} & 5.7450   \\
    &  +$\mathcal{L}_{percep}^{local}$      & \tunder{30.21}   & 0.9311  & \tbold{0.0652}  & \tbold{4.3113}  & \tunder{28.54} & 0.9009 & \tbold{0.0865} & \tbold{3.8116}   \\
    &  +$\mathcal{L}_{percep}^{global}$         & \tbold{30.24}   & \tbold{0.9346}  & 0.0787  & \tunder{4.8536}  & \tbold{28.55} & \tbold{0.9080} & 0.0979 & \tunder{4.3274}  \\
    \arrayrulecolor{gray}
    \hline
    \arrayrulecolor{black}
    Restormer      & \,\,\,\,\,\,\,\,\,\, -                & 29.61   & 0.9277  & 0.1154  & 5.4515  & 28.62  & \tunder{0.9109}  & 0.1225  & 4.9703  \\
    &  +$\mathcal{L}_{percep}^{vgg}$          & 29.59   & 0.9193  & \tunder{0.0639}  & 5.5767  & 28.18 & 0.8848 & \tunder{0.0848} & 5.4517  \\
    &  +$\mathcal{L}_{percep}^{local}$      & \tbold{30.00}   & \tunder{0.9315}  & \tbold{0.0620}  & \tbold{4.5327}  & \tunder{28.70} & 0.9099 & \tbold{0.0791} & \tbold{3.8731}  \\
    &  +$\mathcal{L}_{percep}^{global}$         & \tbold{30.00}   & \tbold{0.9332}  & 0.0718  & \tunder{4.7452}  & \tbold{28.71}  & \tbold{0.9116} & 0.0887 & \tunder{4.1751}   \\ 
    \arrayrulecolor{gray}
    \hline
    \arrayrulecolor{black}
    NAFNet         & \,\,\,\,\,\,\,\,\,\, -                & 32.40   & 0.9565  & 0.0847  & 5.2740  & 30.18  & 0.9304  & 0.0954 & 4.6582  \\
    &  +$\mathcal{L}_{percep}^{vgg}$          & \tunder{32.60}  & \tunder{0.9580}  & \tunder{0.0590}  & 4.9712  & \tbold{30.37}  & \tbold{0.9328} & 0.0761 & 4.2791  \\
    &  +$\mathcal{L}_{percep}^{local}$      & \tbold{32.64}   & \tbold{0.9582} & \tbold{0.0557}  & \tunder{4.9093}  & \tbold{30.37}  & \tunder{0.9322} & \tbold{0.0722} & \tunder{4.1120}  \\
    &  +$\mathcal{L}_{percep}^{global}$         & 32.48   & 0.9568 & \tbold{0.0557}  & \tbold{4.7807}  & \tunder{30.19}  & 0.9279 & \tunder{0.0738} & \tbold{4.0281}  \\ 
    \end{tabular}%
    }
\end{center}
\end{table}

\subsection{Image Deraining}

\begin{table}[htbp]
\caption{Results on the Rain100H dataset \cite{yang2017deep} for single image derain task. The \tbold{bold} numbers indicate the best results and the second bests are marked by \tunder{underlines}.}
\label{tab:image rain}
\footnotesize
\begin{center}
    \renewcommand{\arraystretch}{1.0}
    \setlength\tabcolsep{8pt}{
    \begin{tabular}{l|l|c|c}
    Method & Loss & PSNR & SSIM \\
    \hline
    Uformer & baseline & 30.92 & 0.9028 \\
            & +$\mathcal{L}_{percep}^{local}$ & \tbold{31.30} & \tbold{0.9093} \\
            & +$\mathcal{L}_{percep}^{global}$  & \tunder{30.96} & \tunder{0.9030} \\
    \arrayrulecolor{gray}
    \hline
    \arrayrulecolor{black}
    DRSformer & baseline & 31.25 & 0.9101 \\
              & +$\mathcal{L}_{percep}^{local}$ & \tunder{31.27} & \tbold{0.9145} \\
              & +$\mathcal{L}_{percep}^{global}$ & \tbold{31.29} & \tunder{0.9106} \\
    \arrayrulecolor{gray}
    \hline
    \arrayrulecolor{black}    
    Restormer & normal & 31.54 & 0.9165 \\
              & +$\mathcal{L}_{percep}^{local}$ & \tunder{31.70} & \tbold{0.9185} \\
              & +$\mathcal{L}_{percep}^{global}$ & \tbold{31.87} & \tunder{0.9175} \\
    \end{tabular}
    }
    \vspace{-10pt}
\end{center}
\end{table}

To further verify the effectiveness of our proposed perceptual losses, we applied our method to the task of image deraining. For this task, we chose three SoTA learning-based methods as baselines: Uformer, Restormer, and DRSformer \cite{chen2023learning}. These models are among the most representative and competitive in current literature. We conducted our evaluation using the Rain100H dataset, which includes 1,800 rainy images for training and 100 images for testing. In line with standard practices in deraining research \cite{jiang2021multi, zheng2020single, chang2023direction, wang2023rcdnet}, we assessed the performance using PSNR and SSIM metrics calculated in the Y channel of the YCbCr space. The qualitative results of these experiments are tabulated in Tab. \ref{tab:image rain}.

As expected, models equipped with our proposed loss functions achieved the highest PSNR/SSIM scores. Specifically, the Uformer model registered a gain of 0.42 dB in terms of PSNR. The outcome underscores the robustness of our proposed loss and its applicability to other image restoration tasks.

\subsection{Ablation Study}

In this section, we first study which type of pretrained ViT is more effective in guiding the image deblurring tasks (Sec. \ref{sec: types of vit}). Then, we analyze the effects of different feature representations in ViT in Sec. \ref{sec: features of vit}. Finally, we discuss the setting of key hyperparameters (\ie, layer number of feature extraction and mask ratio) in Sec. \ref{sec: encode of mae}. For convenience, all ablation studies are carried out with the DPDNet and regularized with local MAE perceptual loss $\mathcal{L}_{percep}^{local}$ on the dual-pixel defocus deblurring task.

\subsubsection{MAE, DINO, and ViT}
\label{sec: types of vit}

\definecolor{baselinecolor}{gray}{.9}
\newcommand{\baseline}[1]{\cellcolor{baselinecolor}{#1}}

\begin{figure*}[htbp]
    \centering
    \includegraphics[width=18cm]{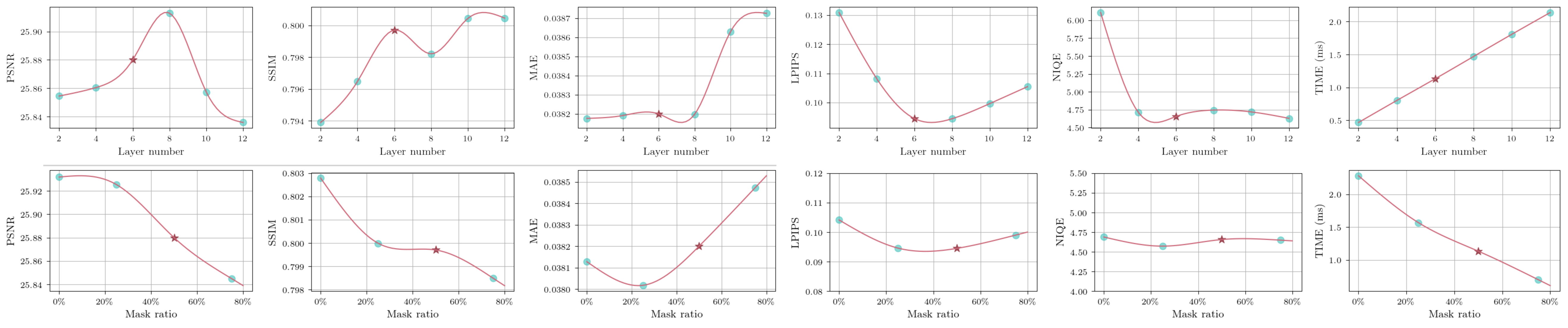}
    \caption{Effect of parameter settings of MAE encoder. We report the quantitative comparisons and time-consuming when features are extracted according to different layer numbers (first row) and mask ratios  (second row) . Default setting is marked as $\color{carmine}\star$.}
    \label{fig:mae encoder}
    \vspace{-5mm}
\end{figure*}

\begin{figure}[htbp]
\footnotesize
    \centering
    \begin{overpic}[width=8.5cm]{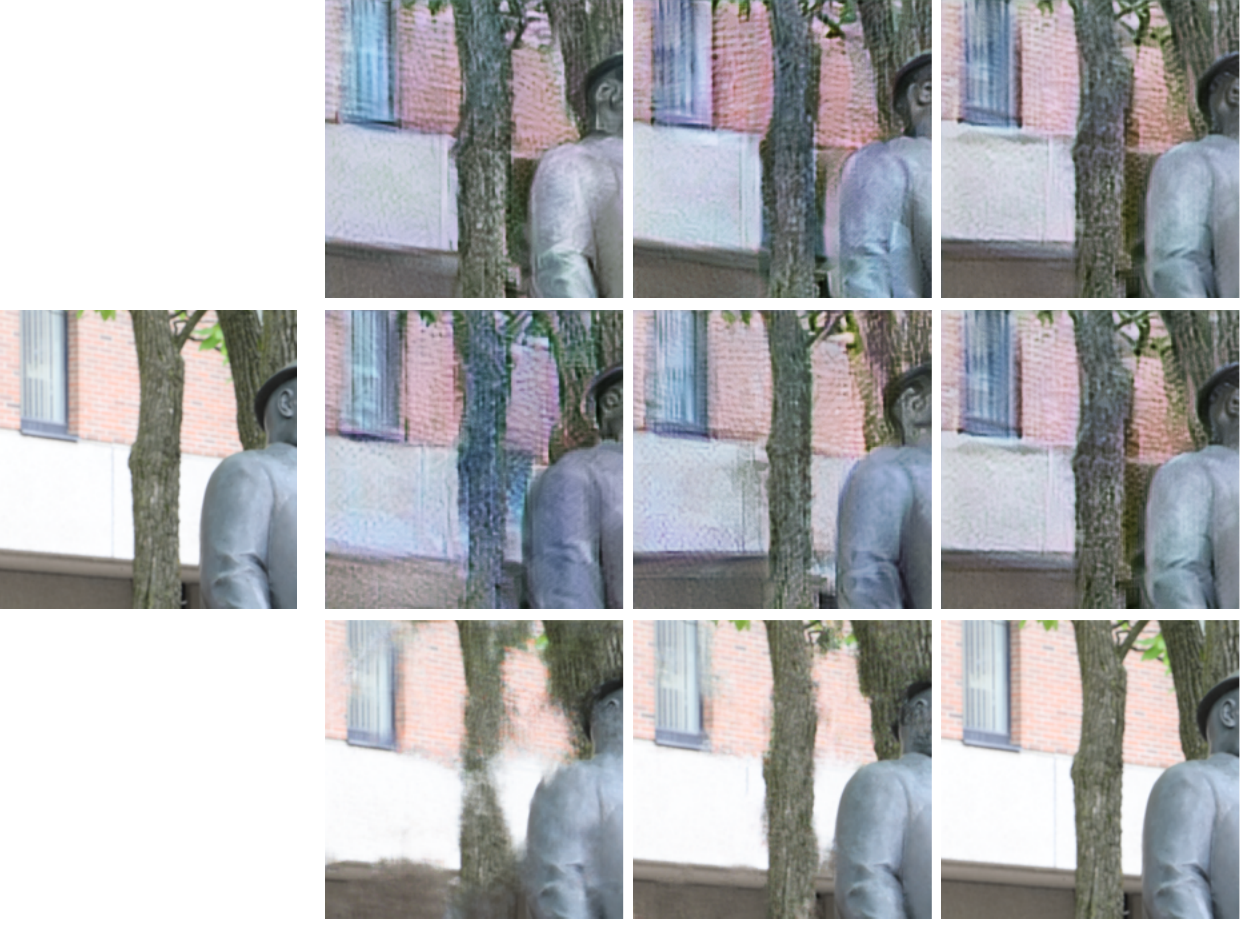}
    \put(7,  25){Source}  % 50,50
    \put(32, -0.3){(a) 25\%} 
    \put(57, -0.3){(b) 50\%}
    \put(82, -0.3){(c) 100\%}
    \end{overpic}
    \caption{The visualization of feature inversion using VGG and MAE models. The first two rows are derived from VGG, while the last row is generated using MAE. From images (a) to (c), we progressively increase the percentage of extracted features used for visualization. The percentage (\eg, 25\%) indicates the proportion of features preserved for inversion.}
    \label{fig: vgg inv}
    %\vspace{2mm}
\end{figure}
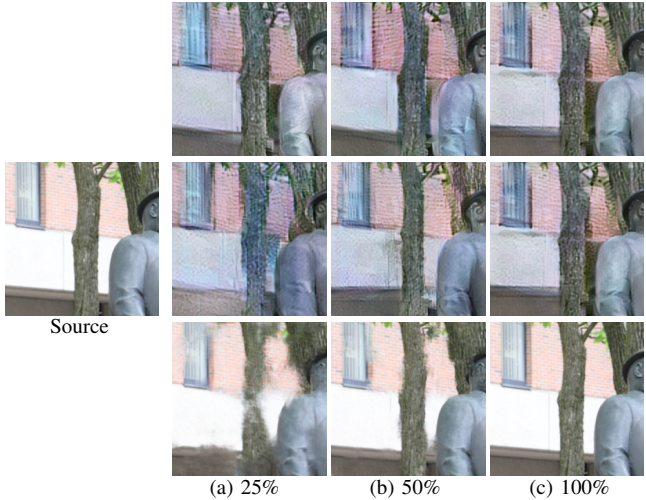

% \color{burgundy}

\begin{table}[htbp]
\footnotesize
\caption{ Evaluation of the impact of the types of pretrained ViTs. We notate the pretrained vanilla vision transformer based on supervised learning as ViT, and simplify the DINO-ViT as DINO. The reported baseline originates from the DPDNet trained on the dual-pixel DPDD dataset. The default settings are marked in \colorbox{baselinecolor}{gray}. }
\label{tab: types of ViT}
\begin{center}
    \renewcommand{\arraystretch}{1.0}
    \setlength\tabcolsep{3.5pt}{
    \begin{tabular}{@{\,\,}>{\raggedright}p{0.073\textwidth}|p{0.050\textwidth}p{0.053\textwidth}p{0.053\textwidth}p{0.053\textwidth}p{0.053\textwidth}@{\,\,}}
    Type  &PSNR & SSIM & MAE & LPIPS & NIQE \\
    \hline
    Baseline  & 25.45 & \tunder{0.7987} & 0.0391  & 0.1531 & 4.715 \\
    ViT       & \tunder{25.61} &0.7967 & \tunder{0.0389} & \tunder{0.1237} & 4.675 \\
    DINO      &  25.57 & \tunder{0.7987} & 0.0390 & 0.1241 & \tunder{4.667} \\
    MAE       &  \baseline{\tbold{25.88}} & \baseline{\tbold{0.7997}} & \baseline{\tbold{0.0382}} & \baseline{\tbold{0.0945}} & \baseline{\tbold{4.656}} \\
    \end{tabular}%
    }
    \vspace{-12pt}
\end{center}
\end{table}

In this section, we compare three representative ViT architectures on the DPDD dataset. The three pretrained ViTs contain almost the same architecture but quite different parameters. The first type is the vanilla vision transformer obtained from supervised learning, which is marked as ViT \cite{dosovitskiy2020vit}. The second and third types, called DINO \cite{caron2021emerging} and MAE \cite{MaskedAutoencoders2021}, respectively, are based on self-supervised learning frameworks. The comparison results are summarized in Tab. \ref{tab: types of ViT}.

The results in Tab. \ref{tab: types of ViT} show that the MAE outperforms other types of ViTs on all metrics. Specifically, compared to the baseline in term of PSNR, the MAE achieves a marginal improvement of 0.43 dB in term of PSNR, while DINO and vanilla ViT demonstrate similar slight performance gains (approximately 0.12 dB $\sim$ 0.16 dB). In addition to PSNR, the LPIPS score of MAE also outperforms the counterparts by a large margin, indicating that the feature representations of MAE facilitate the deblurring model to produce images of high perceptual quality. Another intriguing observation is that ViTs trained in self-supervised learning way perform consistently improvements under both quantitative and perception measurement. In the following ablation studies, we set the MAE to the default setting due to its impressive performance.

\subsubsection{Feature Representation}
\label{sec: features of vit}

\begin{table}[htbp]
\footnotesize
\caption{ Evaluation of the impact of the types of feature representations. The reported baseline originates from the DPDNet trained on the dual-pixel DPDD dataset. The default settings are marked in \colorbox{baselinecolor}{gray}.
}
\label{tab: features of ViT}
\begin{center}
    \renewcommand{\arraystretch}{1.0}
    \setlength\tabcolsep{3.5pt}{
    \begin{tabular}{@{\,\,}>{\raggedright}p{0.073\textwidth}|p{0.050\textwidth}p{0.053\textwidth}p{0.053\textwidth}p{0.053\textwidth}p{0.053\textwidth}@{\,\,}}
    Type  &PSNR & SSIM & MAE & LPIPS & NIQE \\
    \hline
    Baseline  & 25.45 & \tunder{0.7987} & 0.0391  & 0.1531 & 4.715 \\
    Key       & 25.84 &0.7956 & \tunder{0.0381} & 0.0955 & \tbold{4.626} \\
    Query     &  \tunder{25.90} & 0.7966 & \tbold{0.0379} & \tbold{0.0942} & 4.756 \\
    Value     &  \tbold{25.96} & 0.7986 & \tbold{0.0379} & 0.0983 & 4.731 \\
    Token     &  \baseline{25.88} & \baseline{\tbold{0.7997}} & \baseline{0.0382} & \baseline{\tunder{0.0945}} & \baseline{\tunder{4.656}} \\
    \end{tabular}%
    }
    \vspace{-10pt}
\end{center}
\end{table}

In the previous ablation study, we have analyzed the benefits of pretrained parameters among different types of ViT. Here, we further study which types of feature representations (\emph{i.e.}, key, query, value, and token) are more efficient in Sec. \ref{preliminaries}. %The comparison results are reported in Tab. \ref{tab: features of ViT}.
As present in Tab. \ref{tab: features of ViT}, it is obvious that no one type of feature representation has absolute advantages over their counterparts in all metrics. When the query/value features are used for the guidance, the obtained PSNR and MAE scores significantly go beyond those of the baseline. However, the NIQE scores suggest that using the query/value features performs worse than the baseline.
Therefore, it is hard to determine which type of feature is beneficial or useless. The key features suffer from the similar problem that the model taking key features as regularization term shows a significant improvement of perceptual quality while performing poorly measured by quantitative metrics (\eg, SSIM). Considering the balance between perceptual quality and quantitative scores, we employ token features as the default feature representations for our perceptual losses. It is worth noting that the model trained with token features demonstrates comprehensive performance gains compared to the baseline.

\subsubsection{MAE Encoder}
\label{sec: encode of mae}

The ablation studies conducted in this section excel in investigating the impact of key hyperparameter choices in MAE. Here, we analyze two key factors: 1) layer number; 2) mask ratio. For the layer number, we have discussed in Sec. \ref{preliminaries}, which refers to the $l$-th transformer layer used to extract features. The mask ratio indicates that a portion of tokens is randomly masking in the input. 
Visual comparison results are shown in Fig. \ref{fig:mae encoder}.

\begin{table}[htbp]
\footnotesize
\caption{ Evaluation of the impact of the layer number in pretrained MAE. The default settings are marked in \colorbox{baselinecolor}{gray}. }
\label{tab: multiple layers}
\begin{center}
    \renewcommand{\arraystretch}{1}
    \setlength\tabcolsep{5.5pt}{
    \begin{tabular}{@{\,\,}>{\raggedright}p{0.073\textwidth}|p{0.050\textwidth}p{0.053\textwidth}p{0.053\textwidth}p{0.053\textwidth}p{0.053\textwidth}@{\,\,}}
    Type  &PSNR & SSIM & MAE & LPIPS & NIQE \\
    \hline
    Baseline  & 25.45 & {0.7987} & 0.0391  & 0.1531 & \tunder{4.715} \\
    Combine & \tbold{25.99} & \tbold{0.8058} & \tbold{0.0376} & \tunder{0.1120} & 4.814 \\
    MAE       &  \baseline{\tunder{25.88}} & \baseline{\tunder{0.7997}} & \baseline{{0.0382}} & \baseline{\tbold{0.0945}} & \baseline{\tbold{4.656}} \\
    \end{tabular}%
    }
\end{center}
\end{table}

\noindent \textbf{Layer number.} To intuitively illustrate the impact of layer number of the pretrained features, we present a smooth curve to fit the discrete measurement scores of different layers, as shown in Fig. \ref{fig:mae encoder}. We observe that the image quality scores do not vary monotonically as the layer number increases. The optimal results always appear in the middle layer instead of in the beginning or end layer. Another intriguing observation is that the middle layers, \eg, 6th and 8th, tend to produce results with both a better quantitative and perceptual quality. Therefore, it can be concluded that the features extracted from the middle layers of the MAE play a more important role in guiding the deblurring task. Furthermore, for the running time, it is clear that the time cost is linear with the layer depth, that is, the deeper the layer, the more time is consumed.

To further explore the impact of layer numbers in the MAE model, we conducted an ablation study that combines features from multiple layers. The results of this study are presented in Tab. \ref{tab: multiple layers}. We found that combining features from different layers leads to a significant improvement in PSNR and SSIM metrics. However, this approach shows weaker performance in terms of perceptual metrics. For perceptual measurement, both LPIPS and NIQE scores decreased compared to the default setting (\ie, using only the 6th layer).

\noindent \textbf{Mask ratio.} Mask ratio is a vital parameter for MAE, and the model with proper mask ratio will obtain robust and powerful features. Here, we analyze the mask ratio to explore its impact on deblurring performance. Fig. \ref{fig:mae encoder} shows the comparison results in the cases where we set the mask ratio to 0\%, 25\%, 50\%, 75\%, respectively. As can be seen, the varying of mask ratio has not obvious impact on the perceptual quality compared to the cases of layer number. This demonstrates that the information of extracted features is redundant for image deblurring to some extent. In addition, for the running time, we observe that time-consuming does not linearly increase as the mask ratio decreases. We set the mask ratio to 50\% as a default setting considering running efficiency and performance improvement.

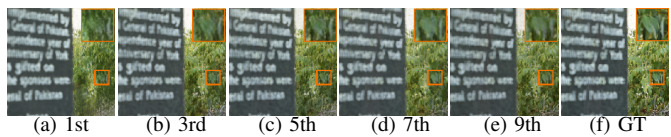
\begin{figure}[htbp]
\footnotesize
    \centering
    \begin{overpic}[width=8.8cm]{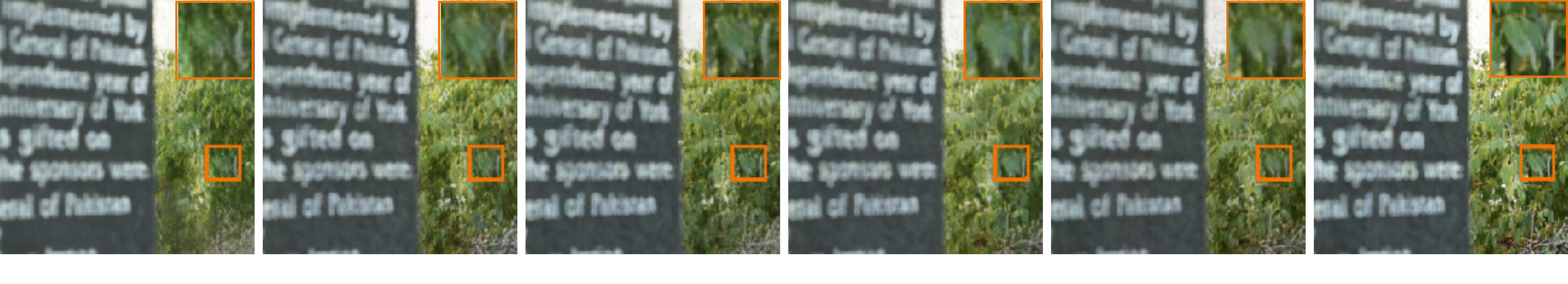}
    \put(4,  -0.420){(a) 1st}  % 50,50
    \put(21, -0.42){(b) 3rd} 
    \put(38, -0.42){(c) 5th}
    \put(55, -0.42){(d) 7th}
    \put(72, -0.42){(e) 9th}
    \put(88, -0.42){(f) GT}
    \end{overpic}
    \caption{ The visualization of feature inversion of different layers in pretrained MAE. }
    \label{fig: localvsglobal}
    \vspace{-5mm}
\end{figure}

\subsection{Visualization}

In this section, we begin by presenting visualizations of feature inversion for both the VGG and MAE models. It serves to demonstrate their respective capabilities in preserving information. Furthermore, we provide additional visualizations to illustrate how local and global semantic information varies with different layer numbers in the MAE model. We use a DIP as the method of feature inversion in the two visualizations \cite{ulyanov2018deep}. Finally, we compare the differences between local and global perceptual losses and discuss their limitations.

\subsubsection{VGG and MAE} 
Fig. \ref{fig: vgg inv} showcases the feature inversion visualization using VGG and MAE models. The first two rows display results from VGG, where we visualize the original input using 100\%, 50\%, and 25\% of the feature maps extracted from the pool5 layer. For feature preservation of VGG, two methods are employed: the first retains features in the channel dimension, as shown in the first row, while the second preserves spatial features, as shown in the second row. The last rows present the results from the MAE model, where we randomly discard tokens, allowing for only one method of selection. It is observed that the feature inversion of MAE closely resembles the source image, even when only 50\% of features are used. In contrast, the VGG results, even with 100\% feature utilization, not only lose some details but also exhibit color artifacts. As the selected features are reduced to 25\%, the MAE still maintains finer details, such as the clear edges of bricks in a wall. However, VGG fails to preserve textural information and produces more inaccurate edges. These observations demonstrate that using MAE as a feature extractor, which preserves more details, is a more robust method than employing the pretrained VGG model.

\subsubsection{Local and Global Semantic Information} 

Fig. \ref{fig: localvsglobal} depicts the visualization results of features from different layers of MAE, achieved through feature inversion. The highlighted regions in the figure clearly demonstrate that the shallower layers of MAE, such as the 1st and 3rd layers, tend to produce images with more pronounced edges and textures. However, these textures are not entirely consistent with the ground truth (GT). Conversely, the deeper layers of MAE, such as the 7th and 9th layers, result in images with coarser contours that appear closer to the GT image, albeit with fewer textures and a somewhat blurred appearance. The visualization suggests that the shallower layers of MAE are more adept at preserving local semantic information, while the deeper layers primarily capture global semantic information.

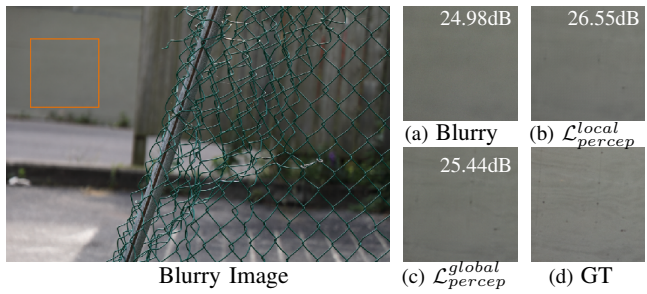
\begin{figure}[htbp]
\footnotesize
    \centering
    \begin{overpic}[width=8.5cm]{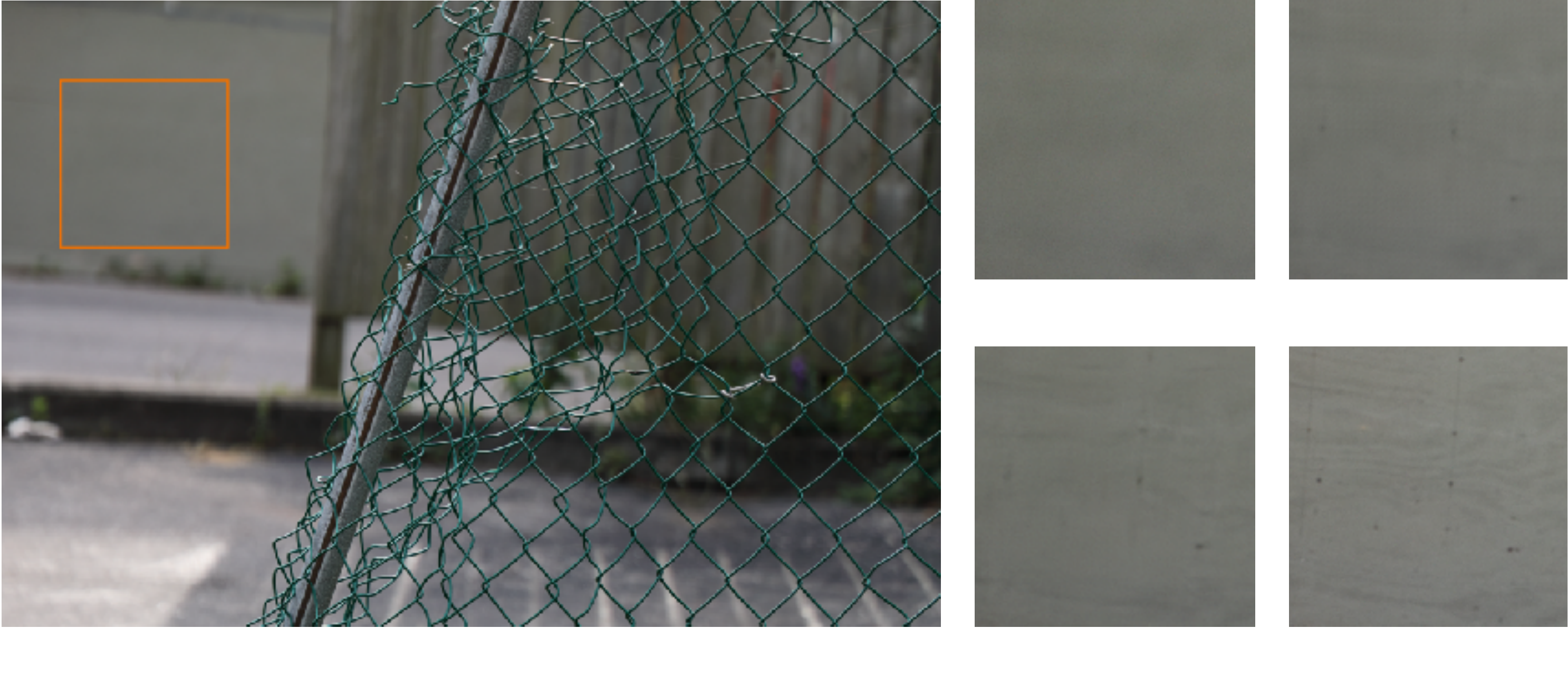}
    % \pgfmathsetmacro{\lastrow}{-1.3 + \vershiftlastfig}
    % \pgfmathsetmacro{\firstrow}{21 + \vershiftlastfig}
    \put(24,  1.1){\small{Blurry Image}}  % 50,50
    \put(62.5, 23.4){(a) \small{Blurry}} 
    \put(82, 23.4){(b) \footnotesize $\mathcal{L}_{percep}^{local}$}
    \put(62, 1.1){(c) \footnotesize $\mathcal{L}_{percep}^{global}$}
    \put(85, 1.1){(d) \small{GT}}

    % \pgfmathsetmacro{\lastrow}{16 + \vershiftlastfig}
    % \pgfmathsetmacro{\firstrow}{39 + \vershiftlastfig}
    \put(68,  41.4){\footnotesize\color{white}24.98dB}
    \put(88,  41.4){\footnotesize\color{white}26.55dB}
    \put(68,  18.4){\footnotesize\color{white}25.44dB}
    \end{overpic}
    \caption{Visual comparisons of the proposed two perceptual losses.To emphasize the differences between these losses, we have cropped and highlighted a flattened region. }
    \label{fig: loss limitations}
    %\vspace{-5mm}
\end{figure}

\subsubsection{Discussion}

In this paper, we propose two types of perceptual losses. Based on the reported experimental results, we can conclude that neither loss achieves the best performance across all metrics on multiple evaluated datasets, as both have their own limitations. To further illustrate their advantages and limitations, we present a representative visual result in Fig. \ref{fig: loss limitations}. The model trained with $\mathcal{L}{\text{percep}}^{\text{global}}$ generates images with richer texture details, which tend to appear more realistic. However, it scores lower in terms of PSNR. In contrast, the model trained with $\mathcal{L}{\text{percep}}^{\text{local}}$ maintains a higher similarity to the original blurred image, displaying fewer edges. Despite this, it achieves the highest PSNR score, outperforming the global model by 1.11 dB.

\section{Conclusion}\label{sec13}

In this paper, we explore harnessing the in-depth transformer properties into image deblurring tasks. To accomplish this, we propose two types of perceptual losses: the local MAE perceptual loss and the global distribution perceptual loss. The former is to directly measure the feature distance in Euclidean space, whereas the latter is defined by comparing the feature distributions. We analyze the effectiveness of proposed perceptual losses on multiple mainstream image deblurring tasks, including dual-pixel defocus deblurring and motion deblurring. Experiments show that the perceptual losses using transformer properties achieve noticeable performance improvement over various baselines and classical perceptual loss based on VGG. 

\ifCLASSOPTIONcaptionsoff
  \newpage
\fi

{
\bibliographystyle{IEEEtran}
\bibliography{sn-bibliography}
}

\end{document}